\documentclass[a4paper,fleqn]{cas-dc}
\usepackage[numbers]{natbib}
\usepackage{comment}
\usepackage{balance}
\usepackage{graphicx}
\usepackage{todonotes}
\usepackage{color}
\usepackage{soul}
\usepackage[super]{nth}
\usepackage{subfig}
\usepackage{amsmath} 
\usepackage{nccmath}
\usepackage[ruled, lined, linesnumbered, commentsnumbered, longend]{algorithm2e}
\usepackage{xcolor}

\def\tsc#1{\csdef{#1}{\textsc{\lowercase{#1}}\xspace}}
\tsc{WGM}
\tsc{QE}
\tsc{EP}
\tsc{PMS}
\tsc{BEC}
\tsc{DE}

 \usepackage{subfig}
 \usepackage{multirow}
\usepackage{array}
\newcolumntype{P}[1]{>{\centering\arraybackslash}p{#1}}
\usepackage{hhline}
\usepackage{colortbl}

\usepackage{soul}
\usepackage{todonotes}

\usepackage{soul}
\usepackage{todonotes}

\begin{document}
\let\WriteBookmarks\relax
\def\floatpagepagefraction{1}
\def\textpagefraction{.001}
\shorttitle{Two-View Fine-grained Classification of Plant Species}
\shortauthors{Voncarlos et~al.}

\title [mode = title]{Two-View Fine-grained Classification of Plant Species}




\author[1]{Voncarlos M. Araújo}

\address[1]{Pontifical Catholic University of Paraná (PUCPR), Curitiba, PR, Brazil}
\address[2]{State University of Ponta Grossa (UEPG), Ponta Grossa, PR, Brazil}
\address[3]{Federal University of Paraná (UFPR), Curitiba, PR, Brazil}
\address[4]{École de Technologie Supérieure (ÉTS), Université du Québec, Montréal, QC, Canada}

\author[1, 2]{ Alceu {S. Britto Jr.}}
\author[3]{ Luiz {S. Oliveira}}
\author[4]{ Alessandro {L. Koerich}}



\begin{abstract}
Automatic plant classification is challenging due to the vast biodiversity of the existing plant species in a fine-grained scenario. Robust deep learning architectures have been used to improve the classification performance in such a fine-grained problem but usually build models that are highly dependent on a large training dataset and are not scalable. This paper proposes a novel method based on a two-view leaf image representation and a hierarchical classification strategy for fine-grained plant species recognition. It uses the botanical taxonomy as a basis for a coarse-to-fine strategy applied to identify the plant genus and species. The two-view representation provides complementary global and local features of leaf images. A deep metric based on Siamese Convolutional Neural Networks is used to reduce the dependence on many training samples and make the method scalable to unknown plant species. The experimental results on two challenging fine-grained datasets of leaf images (i.e., PlantCLEF 2015 and LeafSnap) have shown the proposed method's effectiveness, which achieved recognition accuracy of 0.87 and 0.96, respectively.
\end{abstract}


\maketitle
\section{Introduction}

\begin{figure}
	\centering
		\includegraphics[width=.45\textwidth]{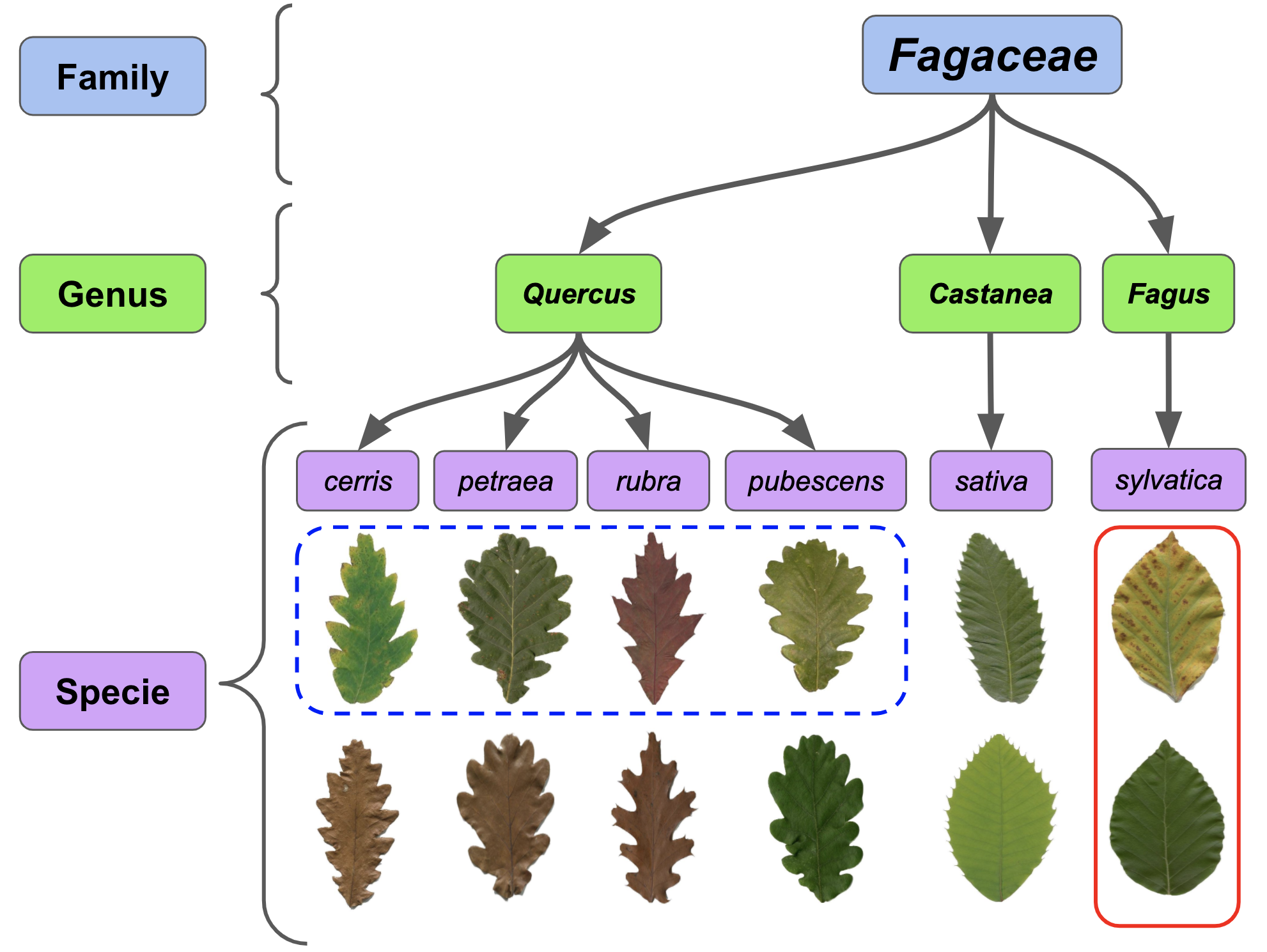}
	\caption{Intra-class and inter-class problem. Inter-class (blue dotted rectangle), has species very similar and intra-class (red line rectangle) contains variations like background, occlusion, pose, color, illumination and plant maturity stages inside the same species.}
	\label{finegrained}
\end{figure}

Automated plant classification concerns the recognition of plant images into botanical species by applying machine learning algorithms \cite{Elhariri2014, Priya2012, Wu2007}. The classification task may be performed on an entire plant's image or just on parts of it, such as branches, flowers, fruits, leaves, or stems. 
This pattern recognition task's main challenge is related to the vast biodiversity of the existing plant species. It is possible to observe the likeness between different species (high inter-class similarity) and sometimes significant differences among samples belonging to the same species (high intra-class variability). The blue dotted rectangle in Figure~\ref{finegrained} shows an example of the possible similarity among different species. In contrast, the red line rectangle presents an example of the difference between samples of the same species caused by shape, color, and texture changes. Such variability is usually caused by the plant maturity or even pose and illumination variation that may result from the image acquisition process. On top of that, there is the unbalancing data problem since some species are scarce in the flora environment and the scalability constraint, as the number of plant species being discovered by scientists is continuously growing.

In the literature, one may find some important strategies to deal with such difficulties inherent to the plant species recognition \cite{Bodhwani2019, Waldchen2018,spatialstructureSiamese,Zhu2019}. However, in the last years, methods based on fine-grained image classification (FGIC) have received special attention from the scientific community \cite{Araujo2018_2, Ge2016,  RejebSfar2015, Sulc2017, mutualchanel, bird11}. Such methods consist of discriminating between classes in a subcategory of objects, such as birds, animals, or vehicle models. Different from the traditional image classification methods, FGIC methods recognize coarse classes firstly. Then, it goes further by discriminating fine classes in which the classification difficulty is greater due to intra-class and inter-class variability like those observed in plant species. FGIC-based methods may explore the taxonomic relationship between the plant classes, which are hierarchically organized based on shared biological characteristics \cite{Schuh2009} into three levels of abstraction: \textit{family}, \textit{genus}, and \textit{species}. Exploring these characteristics may help us distinguish very similar classes by first selecting candidates in a coarse level of the hierarchy, which can be distinguished in the finest level of the hierarchy (coarse-to-fine strategy). The visual recognition of leaves carried out by domain experts is generally based on this hierarchical strategy, the so-called "plant taxonomy relationship" \cite{simpson2010plant}.

With this in mind, we propose in this paper a two-view similarity learning strategy for the fine-grained plant classification, which consists of two stages that exploit different views of the leaf images. In the first stage, a coarse classification by plant genus is carried out using a deep metric based on a Siamese Convolutional Neural Network (S-CNN) to compute the similarity between a testing sample and the reference images previously defined for each plant genus. The deep metric learned from pairs of images provides the distance between two image samples represented by an new plant image and the genus reference images. At this stage, the entire leaf image is used, i.e., the S-CNN model's input is the whole leaf image characterizing a global view in terms of problem representation. The output of this stage is a ranked list of the best genus reference candidates.

In the second stage, a fine classification of plant species is performed. Similarly, an S-CNN is used as a deep metric. The similarity is computed between the cropped test sample and the cropped reference images representing the plant species in the ranking genus candidates list returned by the first stage. Here, a local representation (view) of the leaf image is used, i.e., the S-CNN receives as input a cropped image extracted from the center of the leaf image. The second stage's output is a final ranked list of plant species obtained by combining both stages' results.  

The rationale behind the two-view scheme is to provide different representations of the problem. In the first view, the similarity computed by the S-CNN takes into account global features extracted from the entire leaf image (shape and color). In contrast, in the second view, local features based on texture and the plant veins are considered. Such a representation strategy allows us to treat some specific issues of the leaf classification problem. For instance, species of plants inside the same taxonomy level (e.g., \textit{genus}) may look similar in terms of global features. Still, they present imperceptible tiny local changes in their texture and vein patterns that are important to characterize their species. Using a hierarchical classification decreases the number of classes to be evaluated in a \textit{coarse-to-fine} strategy, reducing the complexity of the plant recognition task.

We carried out extensive experiments and compared the proposed method with state-of-the-art handcrafted approaches and methods based on deep learning using different CNN architectures. AlexNet\cite{NIPS2012_4824}, GoogLeNet\cite{googlenet} and VGG16\cite{SimonyanZ14a} networks were selected as baseline of plant leaf recognition model, and transfer learning was used. For this purpose, a robust experimental protocol was defined based on two challenging fine-grained datasets of plant leaf images: PlantCLEF 2015 \cite{Goeau2014}, and LeafSnap \cite{leafsnap_eccv2012}. In most experiments, the proposed method outperforms several existing methods by achieving superior classification accuracy using few samples to compute the similarity between images. The learned model proposed does not need to be retrained when unknown plant species are added, which makes the proposed method highly scalable.

The contribution of this paper is fourfold: (i) the two-view representation of plant species enables the capture of the leaves' coarse and fine features, which are very useful to distinguish among different \textit{genus} and different \textit{species}, respectively; (ii) a slight comparative evaluation between deep models (AlexNet, GoogLeNet, and VGG16) and deep metric (S-CNN); (iii) the proposed method exploits the natural hierarchy of the problem combining coarse and fine representations into a hierarchical strategy that reduces the complexity of the classification task as a lower number of classes need to be disambiguated at each hierarchical level; (iv) the proposed method is highly scalable as unknown plant species can be easily added using few samples without retraining the S-CNN models. This is highly desirable in plant classification, where the number of new plant species is continuously growing, and the new ones often have few examples for training.

The paper is organized as follows. Section \ref{sec2} reviews the relevant literature in plant classification. In Section \ref{sec3}, the proposed method is described in detail. Section \ref{sec4} presents our experimental findings on plant classification. Finally, Section \ref{sec5} presents our conclusions, future work perspectives, and final remarks.

\section{Related Work}
\label{sec2}

Recently, studies on plant classification based on image processing have become an interesting research topic in computer vision \cite{Araujo2017, Araujo2018_2, Chaki2015, Grinblat2016, Yanikoglu2014}. In the literature, there are many datasets that can be employed to evaluate plant classification methods such as Flavia \cite{Wu2007}, Foliage \cite{Kadir2012}, Swedish \cite{Sderkvist2001ComputerVC}, LeafSnap \cite{leafsnap_eccv2012}, PlantCLEF \cite{Goeau2014}, ICL \cite{Sulc2017}, and MalayaKew \cite{Lee2015}. These datasets represent the problem domain well, exposing the many difficulties such as fine-grained complexity, imbalanced distribution, large intra-class variability, small inter-class variability, and noisy images.

One may find in the literature several contributions to plant species recognition. \citet{Naresh2016} introduced a symbolic approach based on textural features extracted from leaf images for plant species recognition. A modified local binary pattern was proposed to extract features, and the classification was performed using a simple nearest neighbor classifier. Besides, the concept of clustering was used to define multiple class representatives by grouping similar leaf samples using a threshold to create clusters to decrease the intra-class variation. However, in their experiments, they observed the need to incorporate features extracted from other leaf views to improve the recognition accuracy. \citet{Aakif2015} proposed a shape-defining feature, which is combined with morphological and Fourier descriptors. These features were used with artificial neural networks. The method was evaluated on a proprietary dataset of 14 classes and on Flavia and ICL datasets. Their emphasis was more related to the performance in terms of computational time than the recognition accuracy.  

Fine-grained recognition is a challenging problem that consists of recognizing subordinate categories such as species of birds \cite{birdsfine, bird11}, dog breeds \cite{dogsfine} and flower species \cite{flowerfine}. Over the past decade, fine-grained recognition has achieved high-performance levels thanks to the combination of deep learning techniques with large annotated training datasets \cite{Wei2019}. Some recent works have considered deep learning techniques for fine-grained plant classification \cite{Araujo2018_2, Barre2017, Lee2017}. In particular, \citet{Barre2017} and \citet{Lee2017} have shown how convolutional neural networks (CNN) learn representations from plant leaf images using a deconvolutional approach. The most important finding is that shape information alone is not a good choice due to similar leaf contours, especially in closely related species. Therefore, it is important to exploit other kinds of features that may be present in leaf structure. \citet{Araujo2018_2}, explored two types of feature representations of the plant leaf using deep models. The first representation considered the entire image of the plant. After that, the central region of the image is cropped and highlight as the second representation. The approach used a hierarchical classification responsible for combining the outputs of global (entire images) and local (cropped images) features. The features are extracted by the GoogLeNet CNN model pre-trained on the ImageNet dataset. As a result, the authors showed an efficient use of the hierarchical classification instead of a traditional classification. Moreover, the two representations of leaves contributed to the complementarity of features. Despite that, in some cases, the combination of hierarchy levels using product fusion rule was not sufficient to discriminate species with similar characteristics due to not control the number of categories taken to the second level of the hierarchy. The CNN approach also depends on a huge quantity of data to create a robust deep model to handle unbalanced data. They achieved 0.86 using $S$ metric in the final recognition performance for PlantCLEF 2015 dataset employing the data augmentation technique.

As observed in several works, CNN models usually need a high amount of data for training \cite{databig3, databig2, databig1}. For instance, \citet{Barbedo2018} analyzed the impact of the number of training samples on a CNN's accuracy, and he found out that it requires a substantial number of training data to provide solid results. \citet{Barre2017} described an approach based on CNNs for the plant classification, which employs data augmentation based on low-level transformations applied to the leaf images such as shifting, scaling, and rotation. They correctly recognized 86.3\% of the 184 species on the LeafSnap dataset (combining field and lab subset). The resulting CNN model needs to be retrained to include new plant species, which is a time-consuming process. \citet{Barre2017} has also pointed out that most of the misclassified plants belonged to species that show strong visual similarities. {\citet{Song2019} developed a highly discriminative model considering an attention branch-based convolutional neural network (ABCNN) to distinguish between similar leaf features. They achieved 91\% accuracy on the LeafSnap dataset (field subset)}. 
\citet{Zhu2019} introduced a two-way attention hierarchical model using CNNs. The first attention way consists of recognizing the family level based on plant taxonomy. The second attention way is to find a discriminative part of an input image by a heat-map strategy. They conducted experiments in Malayakew and ICL datasets, and the CNN with Xception architecture achieved an accuracy of 99\% in both datasets. They used 90\% of the datasets for training and the 10\% reminding to test. Although the authors stated that they do not use any data augmentation strategy, they have balanced the training dataset. Thus, each class has a roughly equal number of samples. {\citet{Hu2018} conducted experiments on the LeafSnap dataset using a multi-scale fusion convolutional neural network (MSF-CNN). An input image was downsampled to multiple low-resolution images and fed to MSF-CNN. A concatenation operation performs the fusion of features between two different scales. The last layer of the MSF-CNN learns discriminative feature information and aggregates the final feature to predict input plant species. In their experiments, they achieved 85\% accuracy on the LeafSnap dataset considering the field subset.}

A fine-grained classification approach may provide as output just a single class probability or a set of classes so-called "confidence-sets", which include the true class at a given confidence level. To this end, an input image is classified, and the top-$k$ best-ranked classes are selected as the confidence-set. \citet{RejebSfar2015} proposed a hierarchical classification of plants, in which they measured the posterior probabilities for each node of the hierarchy and then created the confidence-set using a confidence threshold. The experiments were carried out on four datasets, where three of them have a balanced number of samples per class in the training set. However, they observed a poor performance on the PlantCLEF 2011 dataset, which is imbalanced, since their strategy fails to recognize the species with few training samples.

\citet{Wang2019} used a few-shot learning method based on Siamese Convolutional Neural Network (S-CNN) to recognize leaf plants. The Euclidean metric was used to measure the distance between features. The structure of GoogLeNet inspired the S-CNN used by \cite{Wang2019}. They evaluated the proposed method on Flavia, Swedish, and LeafSnap datasets. They used a small number of learning samples. The experimental results have shown that the highest classification accuracies (95.32\%, 91.37\%, and 91.75\% for Flavia, Swedish and Leafsnap datasets, respectively.) were achieved using models trained with 20 samples per class. \citet{spatialstructureSiamese} also used S-CNN for plant recognition. They proposed a spatial structure using a deep metric. The S-CNN was used to learn an embedding with similar and dissimilar pairs. Similar pairs were formed using the same plants' organs, and different species of plants organize different pairs. They used the PlantCLEF 2015, and the result was 0.84 using $S$ metric, surpassing all other methods. It is worth mentioning that recurrent neural networks were used to model the spatial structure.
Recently, \citet{Mata2020} proposed a way to learn a similarity metric that discriminates plant species. They compared whether S-CNN models are better than CNN models regarding the performance and computational cost. Also, new species (20 leaves of Costa Rican dataset) never seen by the model S-CNN were evaluated without retraining the proposed model. In their first experiment, they conclude that for datasets with fewer than 20 images per species, the S-CNN performed better than CNN in the context of plant recognition besides the fact of having a lower computational cost. The second experiment has shown that S-CNN can generalize to new plant species without retraining the model.

To the best of our knowledge, from the existing methods in the literature \cite{Barre2017, Ghazi2017,Lee17HGOCNN,Sungbin2015,Song2019,Hu2018,Riaz2020}, only \citet{Wang2019}, \citet{spatialstructureSiamese} and \citet{Mata2020} have exploited deep metrics to compute the similarity between plant images. However, no previous work uses a two-view similarity scheme combined with a fine-grained classification of plants. Plant hierarchy and similarity learning make our method more accurate and scalable, as shown in the next sections.

\section{Proposed Approach}
\label{sec3}
\begin{figure*}
\centering
	\includegraphics[width=.9\textwidth]{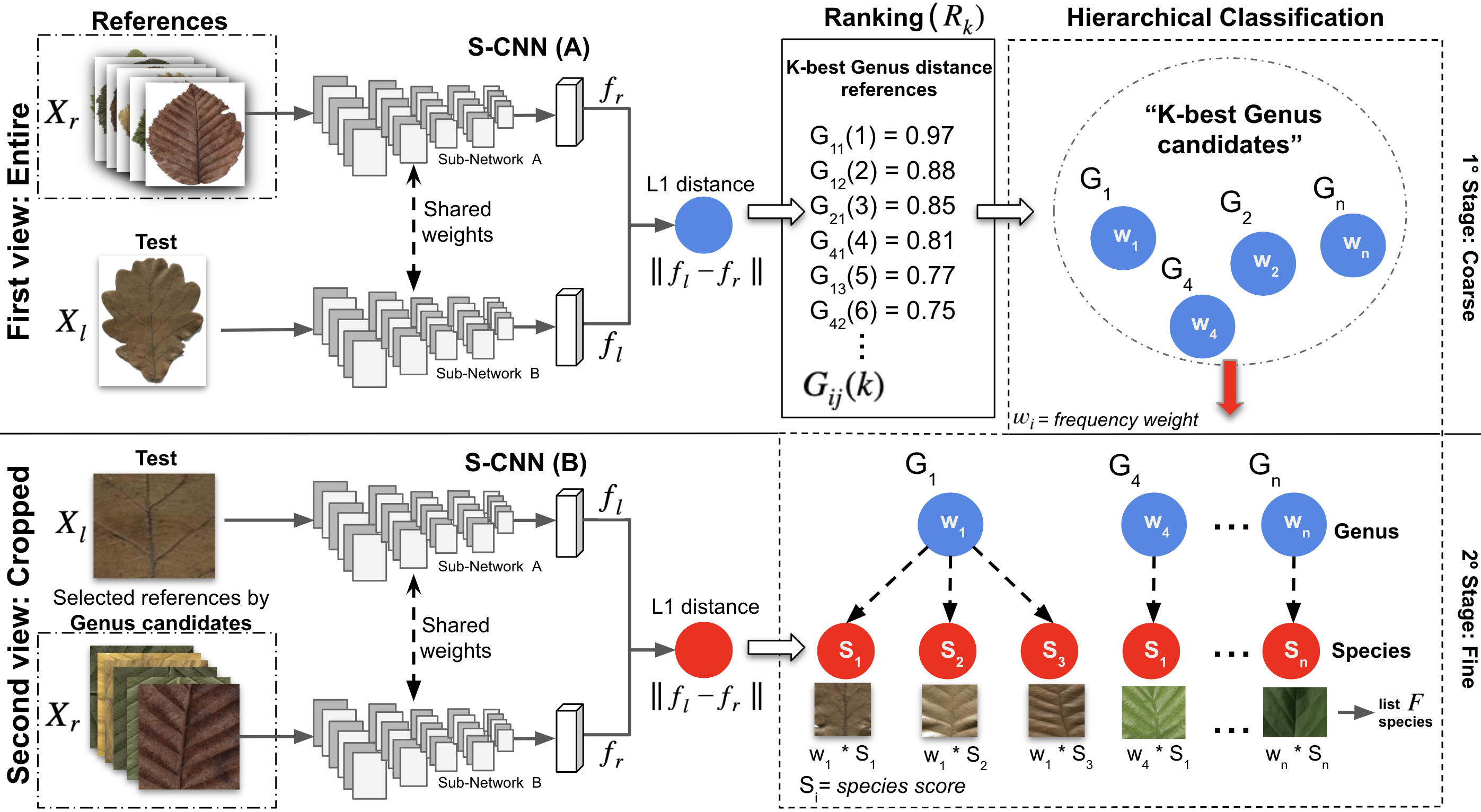}
\caption{Overview of the S-CNN proposed method for fine-grained recognition of plant species from the leaf image.}
\label{SCNNs}
\end{figure*}

We propose a fine-grained approach for the classification of plant species from the leaf image. The coarse-to-fine classification strategy unveils the plant genus in the first stage and then its species in the second stage as illustrated in Figure~\ref{SCNNs}. In the first stage, a coarse classification according to the plant genus is carried out using a deep metric based on a Siamese Convolutional Neural Network (named S-CNN (A)). It computes the similarity between a leaf image $X_l$ and reference images $X_r$ previously chosen to represent each plant genus. Features are extracted by the sub-networks of the S-CNN (A) employing the entire leaf image, which is considered the first view of the proposed approach. The rationale behind that is to compute the similarity between a test image and the genus reference images, assuming a global view of the plant, i.e., representing the leaf by general features such as the leaf shape and color. The first stage's output is the $k$-best genus candidates organized as a ranked genus reference list $R_k$, ordered by the similarity score. We determine the $k$-best genus candidates in the first stage by computing the similarity between the test image and each genus’ reference. Then, we calculate the frequency ($w_i$) of each genus in the list of $k$-ranked candidates ($R_k$). We use $w_i$ to weight the species in the final fusion process.

In the second stage, given the $k$-best genus candidates found in the first stage, a fine classification considering only the plant species which belong to such best candidate genus is performed. Similarly, an S-CNN is used as a deep metric, but now the S-CNN (B) is computed on a different view of the leaf images that consider a local representation (second view). For such an aim, the S-CNN (B) receives as input a cropped image extracted from the center of the leaf image. The idea behind this strategy is to perform a fine classification of plant species based on an adequate representation of the leaf image that focuses on local details such as the texture and the vein patterns that are usually present in the central part of leaves. The second stage's output is a ranked list $F$ of plant species, which is weighted by the first stage's output, as shown in Figure~\ref{SCNNs}. In the next sections, we present the proposed method in detail.

\subsection{S-CNN Learning} 
The similarity between the reference patterns and the leaf image is computed in both representation stages of the proposed method using an S-CNN \citep{Bromley1993}. The difference between the deep metric learning of both stages is that the S-CNNs are trained in different taxonomic groups and views of the leaf image. At the first stage, the S-CNN (A) is trained on the entire leaf image considering the genus group, while in the second stage (S-CNN (B)), only a square region that is taken from the central area of the leaf image is used to train the species group. In other words, we have two siamese nets, S-CNN (A) and S-CNN (B), similarly trained using different views in each taxonomic group (\textit{genus-to-species strategy}).

Both  S-CNNs are composed of two sub-networks that are one of the baselines CNN architectures presented in Section~\ref{secBaselineCNN}. Such twin CNNs are pre-trained on the ImageNet dataset and have shared weights. The original output layer of the pre-trained baseline CNN has 1,000 units that compute the class scores. Such an output layer is replaced by a layer that computes a distance metric between the last fully connected layers of each Siamese twin (Eq.~\ref{eq1_f1}). An image pair ($X_{left}$, $X_{right}$) is the input of the S-CNN (A) and (B). We fine-tune the pre-trained baseline S-CNN model with positive and negative pairs of leaf images, minimizing the loss function denoted in (Eq.~\ref{eqloss}). The positive and negative pairs are generated from each leaf plant dataset samples (see Table~\ref{positiveAndNegative}) to train the S-CNNs how to differentiate the leaf species. An overview of the iterative training algorithm \cite{prototypefewnetwork} used to learn the deep metric models is presented in Algorithm~\ref{alg1}. 

\begin{algorithm}[h]
\caption{Training algorithm.}\label{alg1}
    \SetKwFunction{isOddNumber}{isOddNumber}
    \SetKwInOut{KwIn}{Input}
    \KwIn{ $numEpochs$, training set ($X_T$, $y_T$), $batch\_size$ }
    \For{$i=1$ \KwTo $numEpochs$}{
    	$Batch$ = \textit{get\_batch\_pairs}($X_{T}$, $y_T$, $batch\_size$) \\
      	\For{$j=1$ \KwTo batch\_size}{       
        $X_{left}$, $X_{right}$, $y$ = $Batch(j)$ \\
        $f_l$ = \textit{feature\_extraction} ($X_{left}$) \\
        $f_r$ = \textit{feature\_extraction} ($X_{right}$) \\ 
        $d_\mathrm{w}$ = \textit{L1\_distance}($f_l$, $f_r$) using Eq.~\ref{eq1_f1} \\
        $\delta$ = \textit{compute\_loss} ($d_\mathrm{w}$,\, $y$) using Eq.~\ref{eqloss} \\
        \textit{update\_net\_weights}($\delta$) \\
      	} 
   }
\end{algorithm}

As one can see, for each training epoch, a $Batch$ structure containing image pairs and the corresponding labels ($1$-same class, $0$-otherwise) is created using the image training dataset ($X_T$, $y_T$)  (line 2). Then, the S-CNN extracts the feature vectors $f_l$ and $f_r$ from the images $X_{left}$ and $X_{right}$, respectively. Line 7 computes the $L1$ distance between the extracted feature vectors ($f_l$,$f_r$), as denoted in Eq.~\ref{eq1_f1}, where $M$ is the feature map size. 

\begin{ceqn}
\begin{align}
 \textit{L1} (f_{l}, f_{r})=\sum_{k=0}^{M-1} ||{f_{l, i}-f_{r, i}} ||.
\label{eq1_f1}
\end{align}
\end{ceqn}

According to \cite{distancel}, the $L1$ distance is consistently more preferable than other distance metrics (e.g., Euclidean and cosine distance) for high-dimensional vectors. For instance, when using a VGG16 to compose the S-CNN, we have feature vectors of 4,096 entries. Besides, the work that gave worldwide visibility for Siamese Networks uses the $L1$ distance \cite{Koch2015SiameseNN}.

The computed distance ($d_\mathrm{w}$, in line 7) is the input of the last S-CNN layer in which the loss is computed (line 8), as described in Eq.~\ref{eqloss}.

\begin{equation}
Loss = -[y\log(d_\mathrm{w})+(1-y)\log(1-d_\mathrm{w})].
\label{eqloss}
\end{equation}

\noindent where $y$ is the image pair label (1 or 0). Finally, we update the network parameters (line 9).

\subsection{Pre-trained CNN Baseline} 
\label{secBaselineCNN}

We have selected a pre-trained CNN model to compose our Siamese Neural Network based on a set of experiments described in Section \ref{anali_models_CNN_and_views}. The following CNN architectures were evaluated: AlexNet, GoogLeNet, and VGG16.

\begin{figure}
\centering
\includegraphics[width=.46\textwidth]{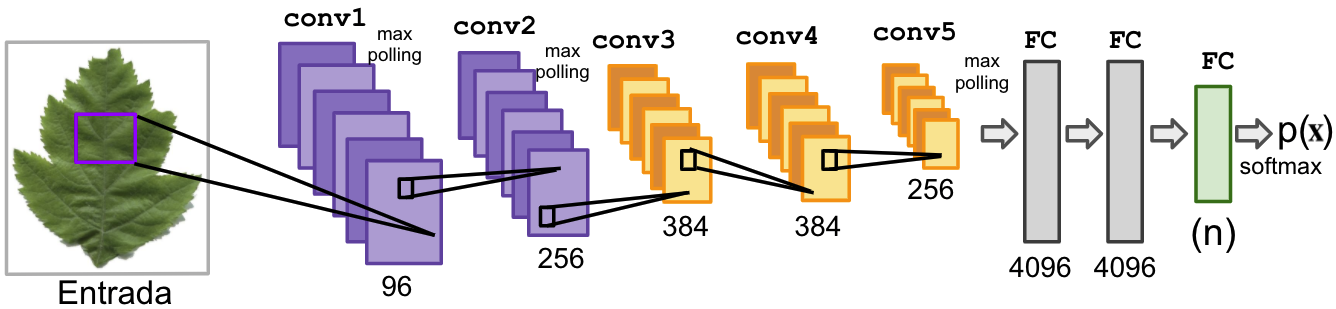}
\caption{The architecture of AlexNet.}
\label{alexnetArchicture}
\end{figure}

Hinton and Alex Krizhevsky created AlexNet. The architecture of AlexNet used in this work is shown in Fig \ref{alexnetArchicture}. The network architecture contains eight weighted layers, the first five being the convolutional layers, and the remains are three (3) fully connected layers. Normalization and pooling layers follow the first two convolutional layers. A single pooling layer follows the last convolutional layer. The third, fourth, and fifth convolutional layers are connected directly. {The fourth, and fifth convolutional layers are unfreeze to fine-tune.} The second fully connected layer is provided to the softmax classifier with the numbers of class labels. ReLU, as the activation function of the first two fully connected layers (fc6, fc7), generates 4,096 values. Finally, the output of the seventh layer of 4,096 data is fully connected to the (n) neurons in the eighth layer (fc8), which (n) represents the number of classes. After training, it (fc8) outputs (n) floating-point values, the predicted result.

\begin{figure}
\centering
\includegraphics[width=.46\textwidth]{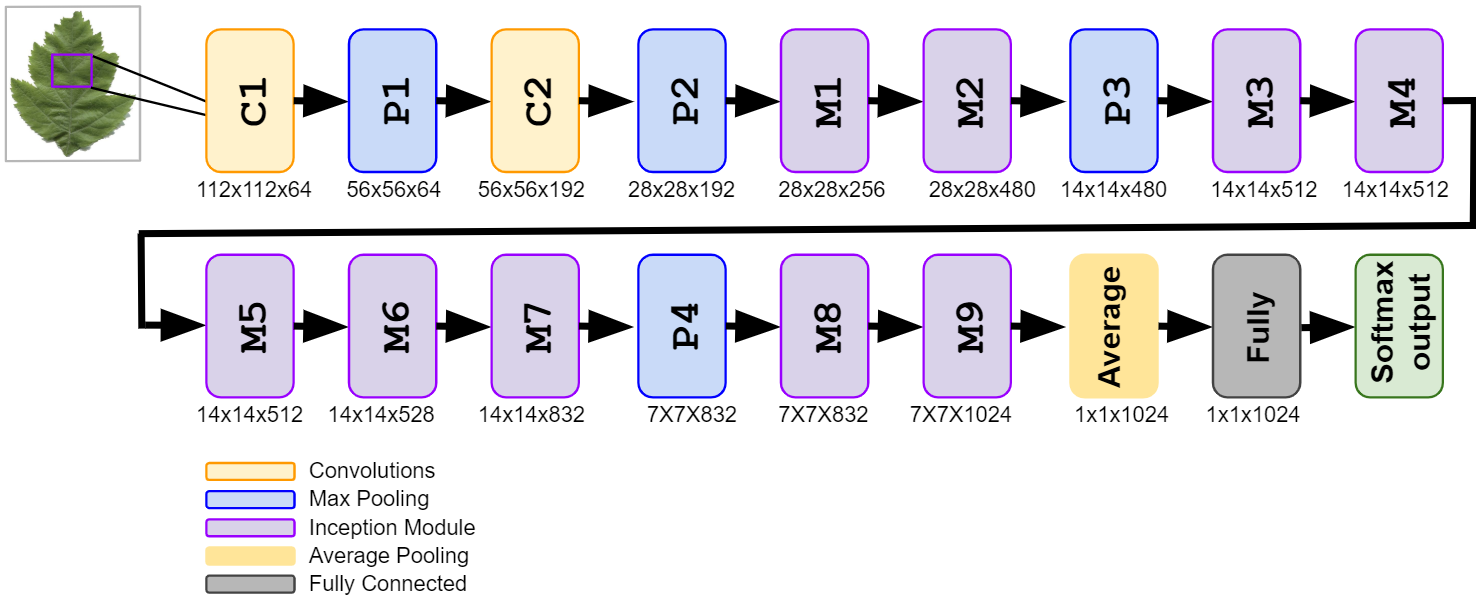}
\caption{The architecture of GoogLeNet.}
\label{googleNetArchicture}
\end{figure}

GoogLeNet was the winner of the ILSVRC 2014 competition, which carried out a top-5 error of 6.67\%. It demonstrated the efficiency to deal with tasks extremely hard for humans, achieving good performance with a low error accuracy. GoogLeNet implements a different way of a network. It incorporates a modern section, which is named as inception module. The inception module uses variable receptive fields, which were created by different kernel sizes. These receptive fields created operations that captured sparse correlation patterns in the new feature map stack. GoogLeNet consisted of 22 layers in total, which was far greater than any network before it. For instance, GoogLeNet drastically reduces the number of parameters, which is only 1/12 of AlexNet. The overall structure of the GoogLeNet network is shown in Fig. \ref{googleNetArchicture}. As shown in Fig.~\ref{googleNetArchicture}, the inception module (M1 until M9) is used on GoogLeNet. {In this paper, the layers before inception M9 were frozen.} The main idea of inception is to localize an optimal local sparse structure and address it as an approximate dense component. The inception module adopts multiple convolutions (1$\times$1, 3$\times$3, 5$\times$5) mixed with the max-pooling layer, which then combines the convolution and pooling results, making GoogLeNet different from a traditional multi-channel convolution. To avoid the heavy computation and overfitting due to a large number of network parameters in the fully connected layers, it directly uses the strategies of averaging pooling and dropout after the inception module, which plays a role in reducing the dimension, as well prevents overfitting to some extent. In this work, there are a total of nine inception models in GoogLeNet architecture. One can find a more detailed overview of this architecture in \cite{googlenet}.

\begin{figure}
\centering
\includegraphics[width=.46\textwidth]{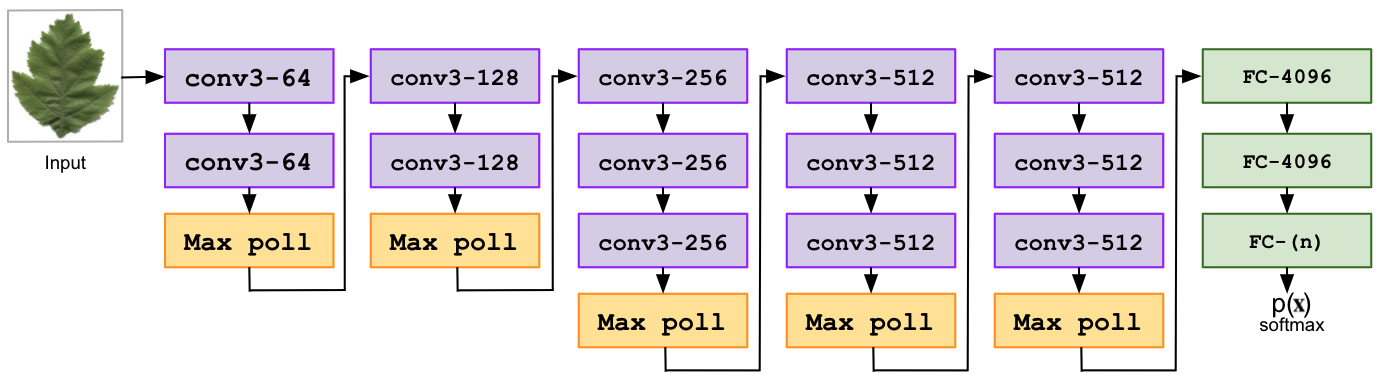}
\caption{The architecture of VGG16.}
\label{VGGArchicture}
\end{figure}

VGG16, developed by Simonyan and Zisserman, was the runner up of the 2014 ILSVRC competition. VGG16 has been upgraded based on AlexNet. The structure of VGG16 is shown in Fig.~\ref{VGGArchicture}. VGG16 contains five blocks of convolutional layers interchanged with five max-pooling layers, followed by two fully connected layers and an output layer. The first two blocks have two convolutional layers with 64 and 128 filters, respectively. The other two blocks have three convolutional layers with 256, 512, and 512 filters, respectively. All filters have size 3$\times$3, and the max-pooling layers have pool size and stride 2. {We fine-tuned the VGG16 unfreezing the last two blocks (Block 4 and Block 5) so that their weights get updated in each epoch during the siamese training.}

\subsection{Hierarchical Classification Strategy} 
A coarse-to-fine classification is performed considering the hierarchical botanic taxonomy of plants. To this end, in the first stage of hierarchical classification schema in Figure~\ref{SCNNs}, the test leaf plant is classified using S-CNN (A) model according to its genus, then its species is defined with S-CNN (B) on the second stage. For each species, we have randomly selected some supervised samples as reference images. The number of reference images per species was experimentally defined (from 1 to 6). The reference images for each genus are those selected for each of its species. A ranked list ($R_k$) of the genus references is the output of the first stage, generating the $k$-best genus candidates of our hierarchical classification corresponding to the coarse classification step, which is based on the global representation of the leaf image. The $k$-best genus candidates are used to select the species to be evaluated in the second stage of the hierarchical classification, which is done considering the leaf image's local representation. Finally, the genus (coarse classification) and species (fine classification) are combined to produce a final ranked list $F$ of the best plant species' hypotheses.

It is worth mentioning that the references for a genus are randomly selected from the training samples but contemplating each of its species. For this purpose, the selection algorithm tries to have at least one reference for representing each species inside the genus. For example, the genus \textit{Prunus} has three (3) species, so the algorithm randomly took two (2) different samples of each species to form the set of six (6) references. It is important to say that the datasets used have a maximum of 6 species per genus. However, even if a genus has more species than the number of references considered in the system, we can see promising results (see our experiments with less than six references in Tables \ref{resultsLifeCLEF} and \ref{resultsLeafSnap}). In most cases, visual characteristics are relatively similar for families and genus in plant species context.

\begin{figure}
\centering
\includegraphics[width=.47\textwidth]{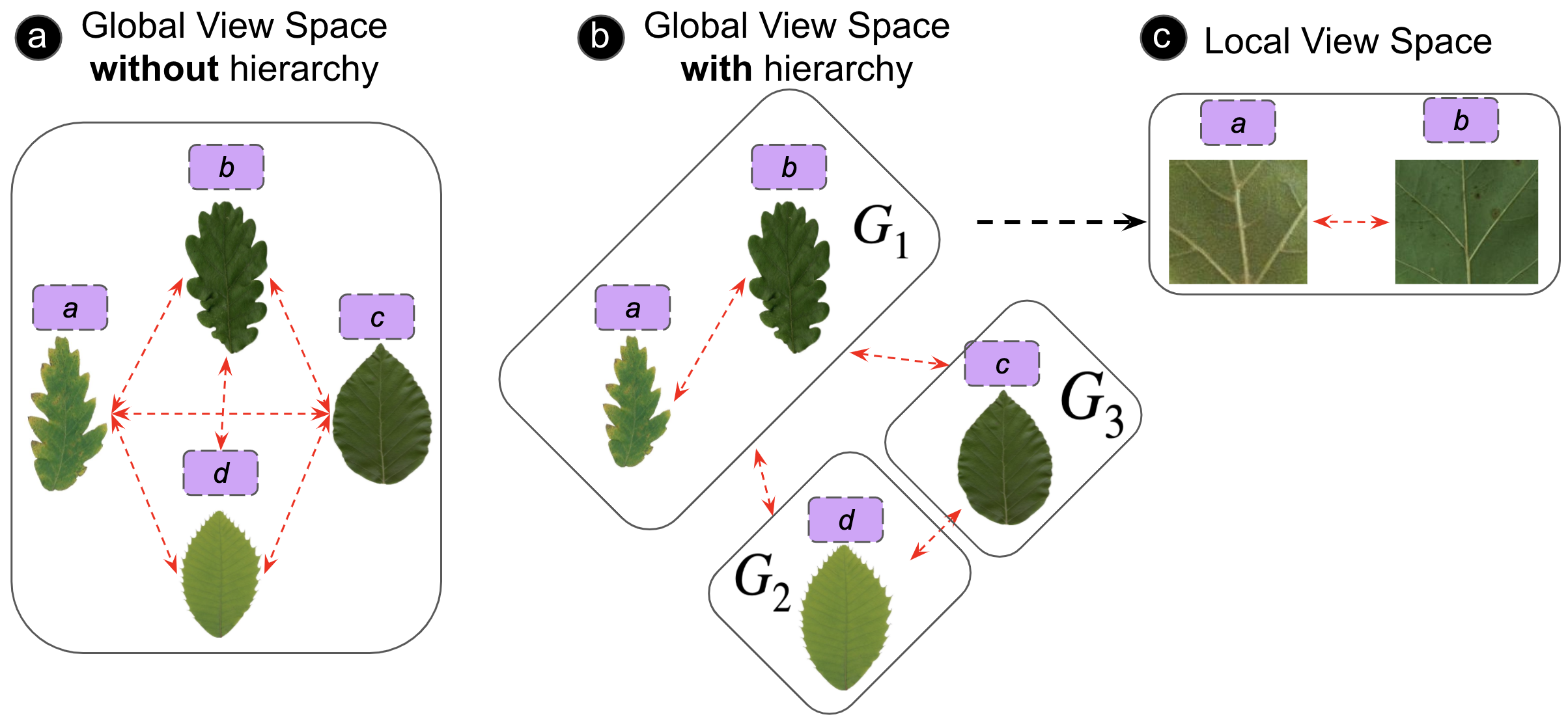}
\caption{a) Different species (a, b, c and d) with similar characteristics are plotted in the global view space; b) The similar categories are separated into three groups that represent the genus taxonomy ($G_1$, $G_2$ and $G_3$); c) Very similar species are discriminated in the local view space.}
\label{distribuitionPlane}
\end{figure}

With this hierarchical classification, we can better deal with the inter-class and intra-class variations observed in the plant species context. For instance, in Figure~\ref{distribuitionPlane}a, four samples of plants are plotted in the global view space without considering the plant hierarchy. In such a scenario, the discrimination must be carried out among all species, making more difficult the discrimination between species with similar characteristics, which increases the complexity of the problem. The hierarchical strategy alleviates the intra-class and inter-class problems, clustering the leaves which have similar characteristics. For instance, by grouping the similar leaves in Figure~\ref{distribuitionPlane}b, we can see that the species belong to three different genus groups ($G_1$, $G_2$ and $G_3$). When the hierarchy is considered (the genus is used), the classification process becomes more manageable, mainly distinguishing species between the genus $G_2$ and $G_3$.

Besides, we observe that some species of the same genus may be alikeness, like those that belong to $G_1$. This motivated us to use an additional representation of leaf (local view) joined with \textit{coarse-to-fine} hierarchical classification in this work. The rationale behind that is to perform some regional analyses of the leaf image (Figure~\ref{distribuitionPlane}c) to deal with such a possible low variability in terms of shape and color between similar species inside some genus ($G_1$). 

\subsection{Fusion Schema} 
{
The final classification is given by the fusion of the outputs of the first and second hierarchical stages. The first stage builds a ranking list $R_k$, containing the genus reference candidates for a given test sample $X_l$ provided by the S-CNN (A), which is trained on the global view representation of the leaf image. The list $R_k$ may be constructed considering different $k$ values, such as: 5, 15, 30, and 50. For instance, if we define the $k = 5$, we can have for a given leaf image a list with the five most similar genus references, with possible repetitions. This ranking list is obtained by comparing the leaf image with the $N_r$ references of each genus ($Nr=6$, for instance). Each $G_{ij}$ in $R_k$ is the $j$-th reference of the genus $i$-th. Finally, we compute the frequency ($w_i$) of each genus in $R_k$.}

{In the second stage, the similarity value $S_{i}$ between the cropped test image ($X_l$) and the reference image ($X_r$) of species inside each genus $G_i$ present in $R_k$ is provided by the S-CNN (B). Now, we consider the local view representation of the leaf image. The second stage's output is a list of species $F$ ordered by the score $\zeta$, computed as described in Eq.~\ref{eqSecondStage}. As one may see, the genus frequency ($w_i$) obtained in the first stage is used to weight the similarity ($S_i$) computed in the second system stage.} 

\begin{ceqn}
\begin{align}
\label{eqSecondStage}
            \zeta = \frac {w_i \cdot S_{i}}{\sum{w_i}}.
\end{align}
\end{ceqn}

With such a fusion schema, we combine global and local views of the leaf image, considering a hierarchical strategy that reduces the number of species in the second stage, decreasing the classification complexity. The global score is the output of SCNN-A, while the local score is the output of SCNN-B. We evaluated different rules (sum, product, and majority voting) to combine the global and local scores produced in the system's first and second stages. The best results were observed when using the schema based on Eq.~\ref{eqSecondStage}. As mentioned, the genus frequency is used as the weight for their respective species. The rationale behind this is that the correct species usually belongs to the genus that most appear in the candidates' list. Thus, we use such a frequency as a weight in the second stage, being important information in the final species decision.

\section{Experimental Results} 
\label{sec4}
This section describes the datasets and experiments used to evaluate the proposed method.

\subsection{Datasets}
We have evaluated the proposed approach on two fine-grained datasets: PlantCLEF 2015 and LeafSnap. {PlantCLEF 2015 is composed of images of various plant structures, such as fruits, flowers, leaves, stems, or even the entire plant. The scope of our work is related to the leaf structure so-called Leafscan. This group contains a total of 12,605 images distributed into 351 distinct species in their training set. The test set contains 221 images with their ground-truth labels, distributed into 60 species. We have used just 6,527 training images to create the image pairs for training the Siamese Network. We reduced the number of training classes to the same number of classes found in the test set (60 classes). It is important to mention that we consider the 351 possible species represented by their corresponding image references during the test. Such a strategy is possible since the Siamese model is composed of pre-trained CNNs, and its output is a similarity value between two images. This allowed us to go further evaluating the scalability of the proposed method. LeafSnap dataset contains 7,719 so-called field images taken by mobile devices from different users in outdoor environments and 23,147 so-called lab images taken of pressed leaves using a high-quality camera with controlled illumination. In this paper, only the field images are applied to evaluate the proposed method}. 

Table \ref{TableDefaultDatasets} presents the original number of classes, training, and testing images available in each dataset. The reason for choosing these two datasets is that both have a wide variability of plant leaf species, representing a challenging scientific task. These datasets' images were gathered by various photographers in globally distributed locations, engaging mixed conditions of background, position, color, and lighting, factors that significantly influence the images' quality. Moreover, both datasets are imbalanced, and for some plant species, there are few samples available for training. Since one of this paper's goals is to classify plant species with a small number of labeled samples, we have evaluated using only six images per class for training the models. It corresponds to the minimum number of training samples per species found in these datasets. Table~\ref{tableTrainingDataset} shows the total number of training images per taxonomic groups (families, genus, and species) when considering only six samples per class in each dataset to be used in the S-CNN proposed method.

\addtolength{\tabcolsep}{-3pt}
\begin{table}[width=.99\linewidth,cols=3,pos=h]
\caption{Original numbers of classes and images for each taxonomic group of the PlantCLEF 2015 and LeafSnap datasets. 
}\label{TableDefaultDatasets}
  \begin{tabular}{ ccccccc}
	\toprule 	
	&   & \multicolumn{4}{c}{Number of} \\
	\cline{3-6}
	Dataset & {Taxonomic}  & Training &   Test &   Training  &   Test  \\
	 & group &   Classes &   Classes  &   Images &    Images   \\
	 \midrule
  {PlantCLEF} & Family 	      		&205		&29 		& 	  &	\\
		 {2015} & Genus	      		&260 	&43				&	12,605			 & 221	\\
  		& Species      		&351		&60				& & 				 \\
  		\midrule

  & Family	      		&35		&35			& 	  &	\\
{LeafSnap}		 & Genus 	      		& 73	& 73				&	7,719  			 & 2,760	\\
  		& Species 	      		&	184	&	184				 & 				 \\
		\bottomrule
		
   \end{tabular}
\end{table}
\addtolength{\tabcolsep}{3pt}

\begin{table}[width=.99\linewidth,cols=3,pos=h]
\caption{The total number of training images per families, genus and species when considering just six samples per class. 
}\label{tableTrainingDataset}
\begin{tabular}{@{}llcc@{}}
\toprule
 & Taxonomic & Number & Total Number\\
Dataset& Group& of Classes & of Images\\ \midrule
\multirow{3}{*}{PlantCLEF 2015} & Family        & 29             & 174            \\
 & Genus        & 43             & 258            \\
 & Species      & 60             & 360            \\
 \midrule
\multirow{3}{*}{LeafSnap} & Family        & 35             & 210            \\
 & Genus             & 73             & 438            \\
 & Species           & 184            & 1,104            \\
\bottomrule
\end{tabular}
\end{table}

We generated training subsets from the training images of Table \ref{tableTrainingDataset}, which have more non-similar pairs than similar ones, as recommended \citet{7899663}. Table~\ref{positiveAndNegative} shows the number of positive and negative samples in our training subsets. A positive sample pair (label 1) means that two images belong to the same category while a negative sample pair (label 0) means that two images belong to different categories.

The PlantCLEF 2015 dataset already has a pre-defined test set made up of 221 leaf images. On the other hand, for the LeafSnap dataset, we randomly choose 15 images per class to compose the test set, totaling 2,760 leaf images. For species with not enough samples, we performed a data augmentation as described in \cite{Araujo2018_2}. 

\begin{table}
[width=.99\linewidth,cols=3,pos=h]
\caption{The number of positive and negative samples of the training subsets used in the proposed S-CNN method.}
\label{positiveAndNegative}
\begin{tabular}{@{}llcc@{}}
\toprule
 & Taxonomic & Positive  & Negative  \\ 
Dataset   & Group     &  Samples &  Samples \\ \midrule
\multirow{3}{*}{PlantCLEF 2015} & Family        & 200           & 300  \\   
  & Genus        & 400           & 600  \\
  & Species     & 800           & 1,200             \\ \midrule
\multirow{3}{*}{LeafSnap} & Family             & 300           & 450             \\
 & Genus             & 600           & 900             \\
 & Species           & 1000          & 1,500              \\
\bottomrule
\end{tabular}
\end{table}

\subsection{Pre-processing}

We use a two-step based pre-processing. First, we remove unwanted structures from the leaf, and then we crop the filtered image. The filtering process originally presented in \cite{Araujo2018_2} is used. First, a copy of the leaf image is converted to gray-scale, followed by a threshold operation using Otsu's method. Afterward, a top-hat technique is applied to remove unwanted objects as the leaf stem. Finally, the bounding box is used to detect just the leaf from the filtered image. {Precise segmentation of the main object in the images avoids unnecessary objects that could interfere with recognition performance \cite{pre1,pre2}}. The final image resolution was 224$\times$224 pixels.

In the second step, the image containing just the leaf is cropped on its center using the algorithm proposed in \cite{augmentor}. We used different cropped image resolutions like 32$\times$32, 64$\times$64, and 128$\times$128. The idea is to evaluate the performance of the S-CNN architecture considering different input resolutions.

\subsection{Analysis and Experiments}
We start this section by presenting four critical analyses that were necessary to define the proposed method. Section~\ref{anali_models_CNN_and_views} evaluates the CNN architectures (AlexNet, GoogLeNet, and VGG16Net) considered alternatives to composing the used S-CNN. Section~\ref{anali2} is used to design the coarse-to-fine hierarchy classification. In Section~\ref{anali2XXX}, the configuration of the two-view hierarchical classification is defined. Following, Section~\ref{anali3}, a fourth analysis shows the importance of the proposed two-view representation of the leaf images.

Additional experiments were performed to evaluate the proposed method. In  Section~\ref{exp1}, we have an overall performance evaluation. Section~\ref{anali1} shows some analysis using the proposed method for plant species recognition. Besides, we evaluated the proposed approach regarding the impact of unbalanced data (Section~\ref{exp2}), scalability, and stability (Section~\ref{exp3}). Finally, in Section~\ref{exp4}, we compare the proposed method with the state-of-the-art.

For the PlantCLEF 2015 dataset, the overall results are computed using the average classification score $S$ proposed in \cite{Goeau2014}. The $S$ metric is defined in Eq.~\ref{eq4} and represents a score related to the rank of the correct species in the list of retrieved species.

\begin{ceqn}
\begin{align}
S=\frac{1}{U}\sum_{u=1}^{U}\frac{1}{P_{u}}  \sum_{p=1}^{P_{u}}\frac{1}{N_{u, p}} \sum_{n=1}^{N_{u, p}}{S_{u, p, n}}\:.
\label{eq4}
\end{align}
\end{ceqn}

\noindent where $U$ is the number of users (who have at least one image in the test data), ($P_{u}$) is the number of individual plants observed by the $u$-th user, ($N_{u, p}$) is the number of pictures taken from the $p$-th plant observed by $u$-th user, and ($S_{u, p, n}$) is the score between 0 and 1 which is equal to the inverse of the rank of the first correct math for the $n$-th pictures taken from the $p$-th plant observed by $u$-th user.

For the LeafSnap dataset, the recognition accuracy is computed by Eq.~\ref{equationLeafSnap}.

\begin{equation}
\label{equationLeafSnap}
            acc = \frac {\text{number of correctly classified samples}}{\text{total number of samples}}\:.
\end{equation}

\subsubsection{\textbf{Evaluating different pre-trained models}}
\label{anali_models_CNN_and_views}

The recognition accuracy of AlexNet, GoogLeNet, and VGG16 is shown in Table \ref{tableArchictures}. The last dense layer of each CNN was adapted to contemplate the number of classes (plant species) of each dataset. Stochastic gradient descent (SGD), with a momentum of 0.9, was used to fine-tuning these deep models. 
Besides, a learning rate of 0.001 was set with a decay of 0.5 at every 512 iterations in a total of 2,048 iterations. Hyper-parameters like batch size and the number of iterations were defined empirically as 32 and 2,048, respectively. We consider the classification of species based on entire images (global view). The experiments were conducted on the PlantCLEF 2015 dataset using the original dataset presented in Table~\ref{TableDefaultDatasets}. For the results on Tables \ref{tableArchictures}, \ref{tableOne} and \ref{tableXXX}, the original training dataset was divided using a randomly stratified strategy into training (70\%) and validation (30\%) subsets. We can observe that VGG16 provided better results under the same experimental conditions, showing an accuracy of 75.09\%.

\begin{table}[width=.99\linewidth,cols=3,pos=h]
\caption{Recognition accuracy (\%) of different architectures considering entire images (just global view).
}\label{tableArchictures}
\begin{tabular}{@{}ccc@{}}


\toprule
 Taxonomic group & Model & Accuracy (\%)\\

\midrule

    &AlexNet    & 71.98      \\   
 Species & GoogLeNet     & 73.30    \\  
    &VGG16   & 75.09  \\
\bottomrule
\end{tabular}
\end{table}

\subsubsection{\textbf{Plant family, genus and species classification}}
\label{anali2}

This section evaluates how to define the coarse and fine levels of the proposed hierarchical strategy. To this end, we performed the classification of leaf images using the pre-trained VGG16 model, taking each taxonomic group (family, genus, and species) individually and considering different views: global and local.  Table~\ref{tableOne} shows the results observed on the PlantCLEF 2015 validation dataset.

\begin{table}[width=.99\linewidth,cols=3,pos=h]
\caption{Individual classification considering each taxonomic group: Family, Genus and Species and different views: global and local with multiple-resolutions: 32$\times$32, 64$\times$64 and 128$\times$128. 
}\label{tableOne}
\begin{tabular}{@{}ccccc@{}}


\toprule
  Taxonomic & \multicolumn{4}{c}{ Accuracy (\%)}
\\ \cline{2-5}

  Group      &Global View  &  & Local View  & \\
\cline{3-5}
      &  &  32$\times$32  & 64$\times$64 & 128$\times$128 \\
\midrule
         Family& 69.51     & 63.01    & 60.47 & 59.16          \\
  Genus& 85.89        & 67.33        & 64.41 & 62.35     \\
         Species& 75.09       & 74.00  & 72.35 & 71.18            \\  \bottomrule
\end{tabular}
\end{table}

\begin{figure*}
\centering
	\includegraphics[width=.9\textwidth]{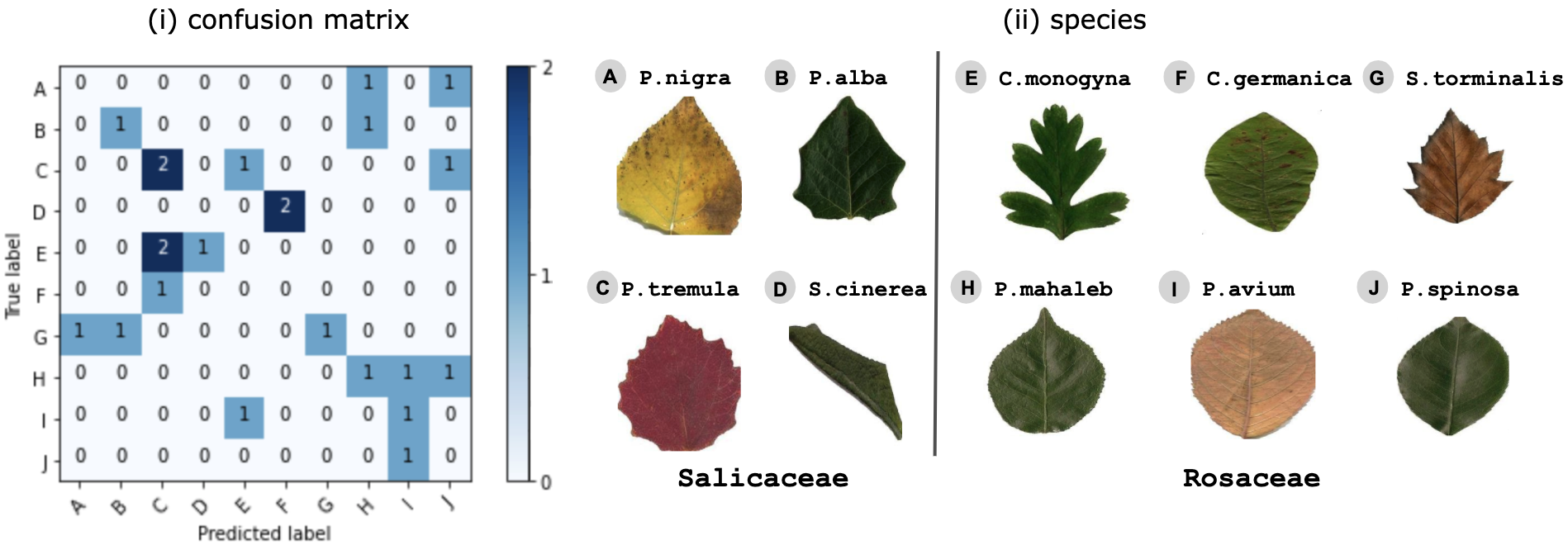}
\caption{Performance of VGG16 trained on Family groups: (i) confusion matrix (ii) Species separated by Families: \textit{Salicaceae} and \textit{Rosaceae}}
\label{Families}
\end{figure*}

\begin{figure*}
\centering
	\includegraphics[width=.9\textwidth]{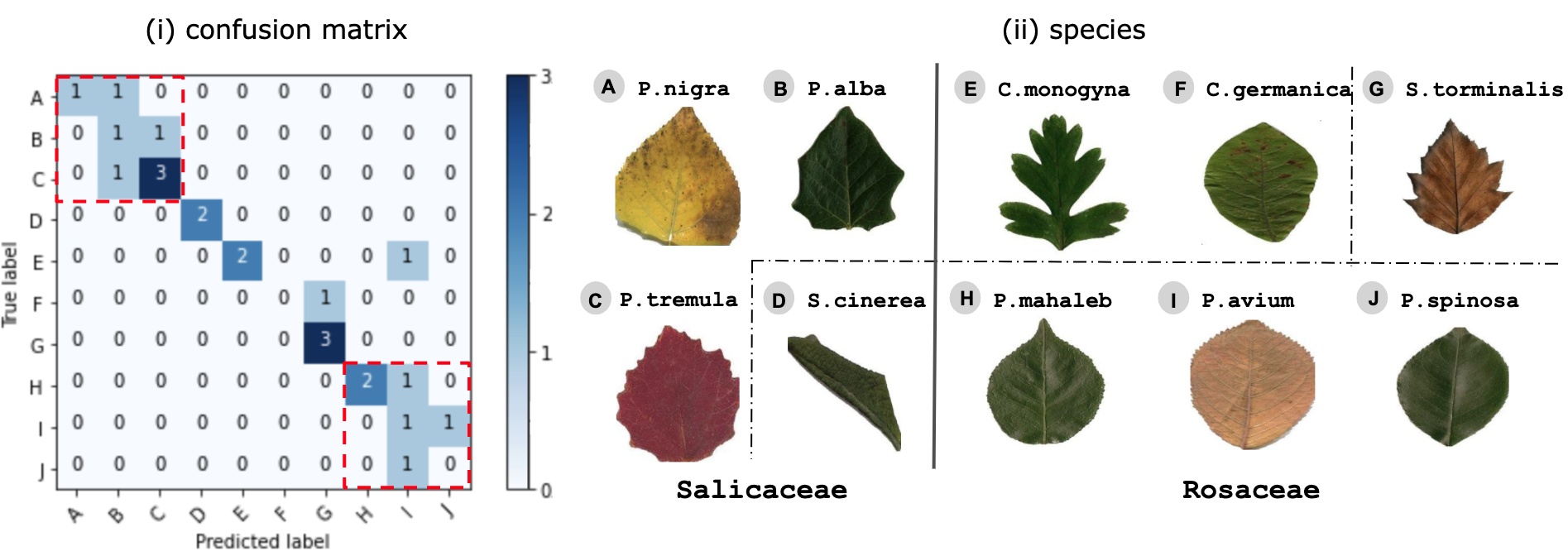}
\caption{Performance of VGG16 trained on Genus groups: (i) confusion matrix (ii) Species separated by Families: \textit{Salicaceae} and \textit{Rosaceae} and Genus: \textit{Populus}, \textit{Salix}, \textit{Crataegus}, \textit{Sorbus} and \textit{Prunus}.}
\label{Genus}
\end{figure*}

As we can see, the local view individually has a worse performance than the global one for all taxonomic groups, indicating that we must avoid the local view in the first level (coarse) of the hierarchy. Table~\ref{tableOne} demonstrates that the VGG16 model is better when using the genus compared to family and species. To better understand the results of plant species recognition per each taxonomic group, confusion matrices are presented in Figs.~\ref{Families} (i) and \ref{Genus} (i). They consider family groups for better visualization and only two plant families (\textit{Salicaceae} and \textit{Rosaceae}) with the corresponding species.
Still, in Table~\ref{tableOne}, we can see that for the local view representation is better the use of cropped images of resolution 32$\times$32 compared to 64$\times$64 and 128$\times$128.

Fig.~\ref{Families} (i) shows the confusion matrix using the families \textit{Salicaceae} and \textit{Rosaceae} of the PlantCLEF 2015 test set. The worse accuracies were observed for the species \textit{S.cinerea}, \textit{C.monogyna}, and \textit{S.torminalis}. We notice that the family model cannot properly classify the leaf species that have different types of features inside the same family. The \textit{S.cinerea} (Fig.~\ref{Families} (D)) provided a false negative error with \textit{C.germanica} (Fig.~\ref{Families} (F)). This confusion may have been caused by the green body color and fine shape, which is quite different from the remaining species inside the \textit{Salicaceae} family. On the other hand, \textit{C.monogyna} (Fig.~\ref{Families} (E)) and \textit{S.torminalis} (Fig.~\ref{Families} (G)) have characteristics dissimilar from leaves inside of \textit{Rosaceae} family, causing misclassification. In any case, we can see that the confusions inside the family group are between species that contain different attributes in terms of color, shape, and features.

Fig.~\ref{Genus} (i) shows a confusion matrix for the same two families \textit{Salicaceae} and \textit{Rosaceae}. Using a VGG16 trained on the plant genus, instead of on the family, we separate the species in genus groups represented by the dotted lines in Fig.~\ref{Genus} (ii). By separating them, we have inside the \textit{Salicaceae} family, two genus (\textit{Populus} and \textit{Salix}) and for \textit{Rosaceae} family, three genus (\textit{Crataegus}, \textit{Sorbus} and \textit{Prunus}) were formed. We observe in Fig.~\ref{Genus} (i) that the VGG16 model trained with genus increases recognition performance due to a better leaf features separability. However, the greatest confusions occur with similar species inside the same genus (highlighted with red dotted lines in the Fig ~\ref{Genus} (i)). Normally, the species inside the same genus have similar visual characteristics, such as the shape. For instance, the three most similar species are \textit{P. nigra} (Fig.~\ref{Genus} (A)), \textit{P.alba} (Fig.~\ref{Genus} (B)) and \textit{P. tremula} (Fig.~\ref{Genus} (C)) represented by the genus named as \textit{Populus}. This similarity is reflected in local confusions since these three species have curved boundaries and the same shape sketch. Despite the similarity between species, the hits in the confusion matrix provided by the VGG16 trained on the genus (Fig.~\ref{Genus} (i)) is better from that considering the VGG16 model trained on family (Fig.~\ref{Families} (i)).

These preliminary experiments indicate that the coarse-to-fine strategy must consider the genus in the first level of the hierarchy strategy instead of family or species. The results are also better using global view in the first level than local one as shown in Table \ref{tableOne}. However, there are still mistakes related to species very similar in appearance (like those presented in red dotted line in Fig.~\ref{Genus}~(i)), that require additional effort to be classified. To deal with this problem, we propose the two-view representation of the leaf image in which global and local features are combined in a hierarchical classification schema.

\subsubsection{\textbf{Configuration of two-view hierarchical classification}}
\label{anali2XXX}

 Table~\ref{tableXXX} shows the use of the taxonomic groups in a diverse pool of hierarchical combinations and which view is used in each level of the hierarchy. For instance, in ID $\#$1, we evaluate the hierarchy classification using Family+Species, which means the Family taxonomic group is used to train the VGG16 model in the first level of the hierarchy, and the Species taxonomy group is trained posteriorly. So, the solution is tested in a \textit{coarse-to-fine} hierarchy way. In the fourth column of Table \ref{tableXXX}, we choose which type of view (global or local) is used in each level of the hierarchy to evaluate diversified combinations. Whether two marks are selected in the fourth column means that the global view is used in the \textit{coarse} level of the hierarchy and local view in the \textit{fine} one. Otherwise, one mark, only the marked view is used in both levels.
 
 \begin{table}[width=.99\linewidth,cols=3,pos=h]
\caption{Diverse combination of \textit{coarse-to-fine} hierarchical classification, according with taxonomic groups and representation of plant leaf views. 
}\label{tableXXX}
\begin{tabular}{@{}cclcccc@{}}


\toprule
Model &  ID & Hierarchical   & \multicolumn{2}{c}{ View } & Accuracy

\\
& \# & combination     &Global   &  Local   & (\%)\\
\midrule
 
      \multirow{6}{*}{VGG16}     &  1  &  Family+Species&       &   $\surd$        &     60.21  \\
                &  2  &  Family+Species&    $\surd$    &         &     62.76  \\
                &  3  &  Family+Species&    $\surd$    &   $\surd$        &     67.55  \\
          &  4  &  Genus+Species&       &   $\surd$        &     64.33  \\
  &    5 & Genus+Species& $\surd$      &           &     78.98  \\

    &    6 & Genus+Species& $\surd$     &   $\surd$ & 83.11    \\

         \bottomrule
         
\end{tabular}
\end{table}

The lowest performances are achieved when the combinations are carried out with the family taxonomic group as the first level of the hierarchical classification (ID $\#$1, ID $\#$2, and ID $\#$3), even though using different views. The individually poor performance for the family taxonomic group achieved in Table \ref{tableOne} makes all combinations that use the family group have the worst accuracy. Because of a hierarchy classification propriety, the error persists in the following classification levels independent of which taxonomy or type of view is used. The ID $\#$4, ID $\#$5, and ID $\#$6 start by using genus in the first level (\textit{coarse}) of the hierarchical classification and species secondly (\textit{fine}). The performance has a soft advance with ID $\#$5 and ID $\#$6. Despite ID $\#$4 remains the poor performance due to the use of local view in both levels of hierarchy, ignoring discriminating characteristics like leaf shape and boundary. ID $\#$5 uses global features for both levels. The errors occur in the second level. Inside the same genus, species with similar characteristics are hard to discriminate using global view only. Finally, ID $\#$6 brings the best result by use two-view in a hierarchical classification strategy.

Assuming the results without hierarchical classification in Table \ref{tableOne}, we have a performance of 75.09\% using species as final output. Using the hierarchical combination (Table \ref{tableXXX}), we increase the accuracy by 3.89 percentage points with ID $\#$5 for the same case (global-view), confirming the assumption that the hierarchy strategy can improve the performance of classification. Besides using ID $\#$6, we increase by 8.02 percentage points, further enhancing the classification by adopting a local view of the plant leaf image in a two-view hierarchical \textit{coarse-to-fine} classification solution.

\begin{figure}
	\includegraphics[width=.48\textwidth]{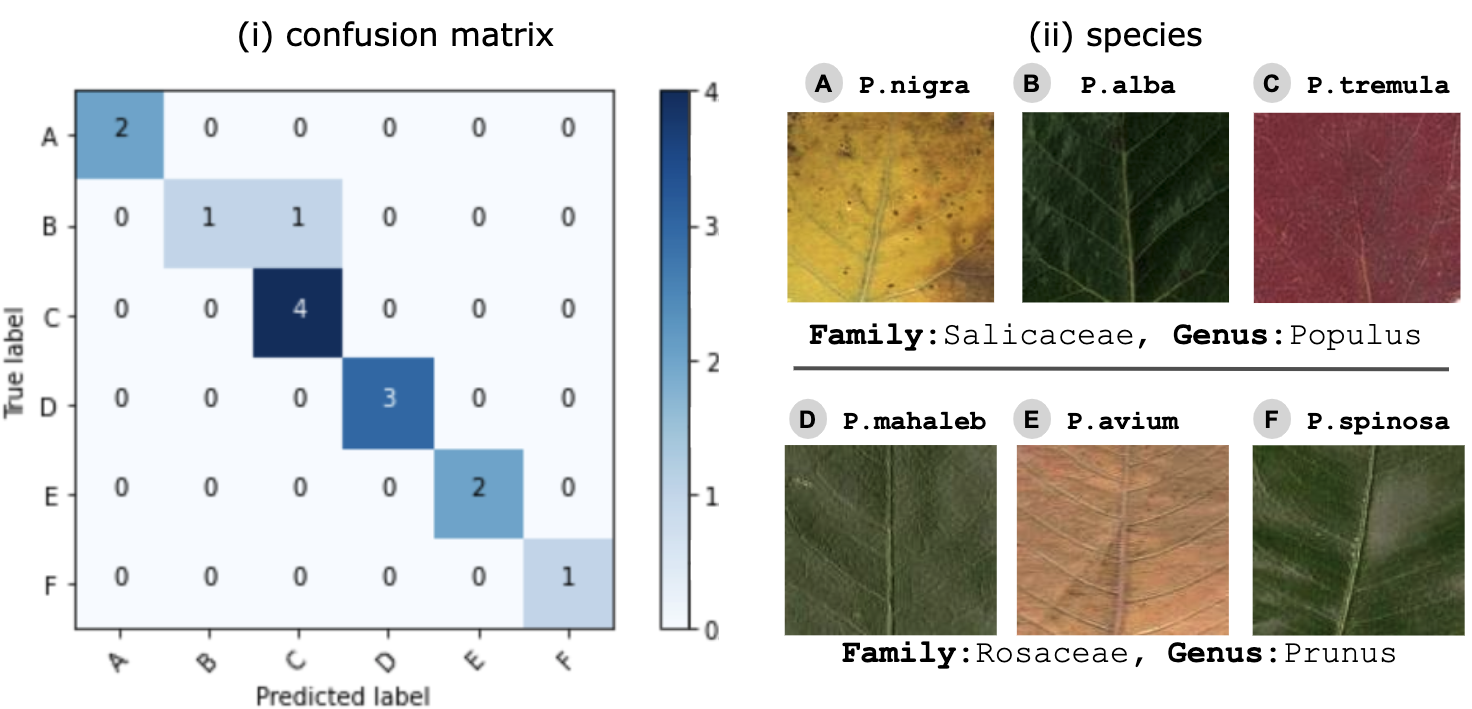}
\caption{Performance of VGG16 trained with local-view representation on Genus groups:(i) confusion matrix (ii) Species separated by Families: \textit{Salicaceae} and \textit{Rosaceae} and Genus: \textit{Populus} and \textit{Prunus}}
\label{local_performance}
\end{figure}

Finally, in Fig.~\ref{local_performance}, we evaluate the impact of using the local view in the second level of the hierarchy. To show the effectiveness, we plot the confusion matrix for specific cases in which the error persists when used just global views. The specific cases have already been presented in Fig.~\ref{Genus} (i) highlighted by dotted red lines and were plotted in a new confusion matrix in Fig.~\ref{local_performance} (i) with a new representation (local view) in the Fig.~\ref{local_performance} (ii). We noticed an increase in the performance results in almost all species when the local view is employed. This occurs because, in terms of global features (general characteristics), the species inside the same genus (red dotted lines in Fig.~\ref{Genus} (ii)) are very similar. However, when plotted in a local view (Fig.~\ref{local_performance} (ii)), they present tiny and discriminate differences among species. The performance
using local view extract features in detail (patterns
veins and texture of leaf plant) able to distinguish similar species. Comparing with ID $\#$5 in Table \ref{tableXXX} that uses just global view in both levels, the results improve by 4.13 percentage points when adopting the local view in the second level of hierarchy (ID $\#$6). In these experiments, we consider the top class returned in the hierarchical classification to pass to another level. Thus, after this previous analysis, we defined our S-CNN proposed approach as genus-to-specie hierarchical classification, using genus as coarse classification and species as the fine classification. Employing global view and local view for the first and second levels of hierarchy respectively as seen in the scheme of Figure~\ref{SCNNs}.

\subsubsection{\textbf{Analysis of the two-view representation}
\label{anali3}}

\begin{figure}
	\includegraphics[width=.48\textwidth]{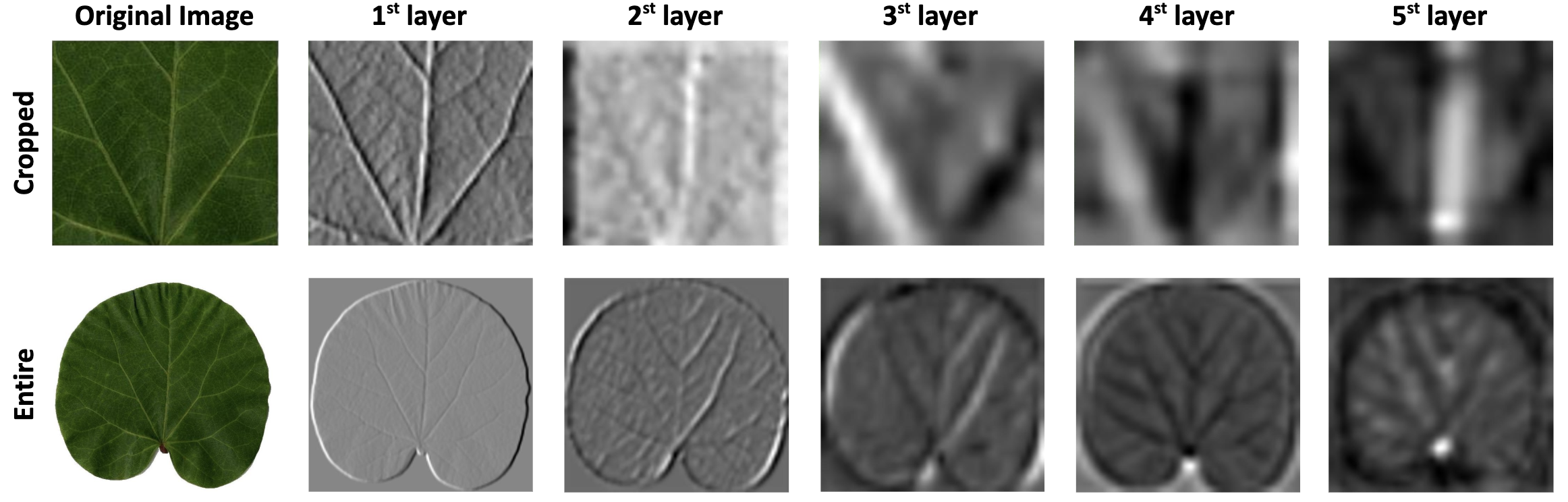}
\caption{Two-view representation - the feature maps along the layers of the S-CNN for entire and cropped images.}
\label{CropAndEntireHighlights}
\end{figure}

As we may see in Table \ref{tableXXX}, the two-view representation (ID \#6 in that table) is a promising alternative to boost the classification performance for the proposed hierarchical classification. With the experiments presented in the last two sections, we gain two essential intuitions regarding leaf features. Firstly, leaf shape alone is not the right choice for identifying plants because of the common occurrence of similar leaf contours, especially in closely related species. In these situations, the venation pattern is a more powerful discriminating feature. Secondly, the strategy of combining global and local features presented in the leaf images is promising. However, what does CNN learns in each view (entire image and cropped image)? To answer this question, we explored the CNN layers of each view in Figure~\ref{CropAndEntireHighlights}.

We observed that, at each CNN layer, similar information is extracted but from different perspectives. Similarly, in the sense that both models provide in the first layers more low-level features, while in the deeper layers, we can observe more class-specific features. Different perspectives since the models have as inputs different views of the plant leaf. For instance, the first convolutional layer tends to extract low-level features like borders. However, it is clear the complementarity between the two networks. The global view provides the whole leaf's edges (shape information), while the local view provides venation-like features. Similar complementarity can be visually observed between each corresponding layer. In general, the entire leaf image (global view) provides features related to the leaf image's general shape and texture. On the other hand, the features extracted from the cropped leaf image (local view) tend to capture inherent local patterns of venation. Such a general-to-specific leaf representation allows us to combine information from the leaf shape and venation, as usually done by plant taxonomists.



\subsubsection{\textbf{S-CNN performance evaluation}}
\label{exp1}

In this section, the proposed S-CNN hierarchical classification based on a two-view similarity scheme is evaluated. Tables~\ref{resultsLifeCLEF} and~\ref{resultsLeafSnap} present the results for PlantCLEF 2015 and LeafSnap datasets, respectively. The best results for both datasets were achieved using six reference samples per class ($N_r$=6) and 30 genus reference candidates in the ranked list $R_k$. In the second stage, the accuracy is computed considering the first result returned in the list $F$ of species (top-$k$=1). We reached a 1.0 accuracy rate with top-$k$=5 for both datasets as reported in Table~\ref{ResultsTopN}.

\begin{figure}
	\includegraphics[width=.48\textwidth]{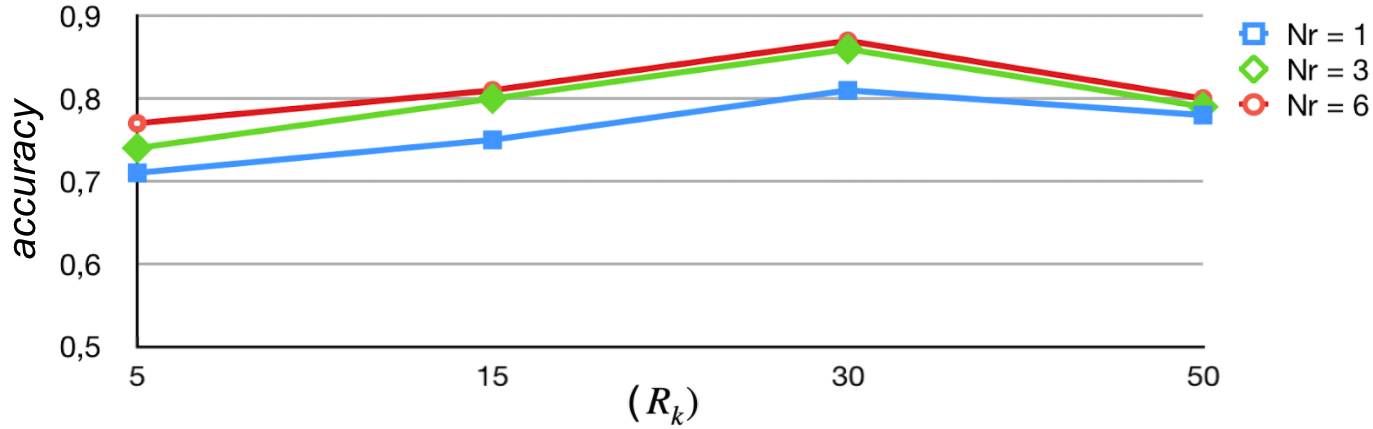}
\caption{Classification accuracy considering different number of references ($N_r$) and sizes for the ranking list of genus reference candidates ($R_k$) in the first stage (\textit{coarse}) of hierarchy using the PlantCLEF 2015 dataset.}
\label{ImageCLEFresults}
\end{figure}

It is important to notice that the number of references $N_r$ has a significant impact on the results. Therefore, we show in Figure~\ref{ImageCLEFresults} different $N_r$ values over different genus references candidates $R_k$. It is expected that as the ranking list $R_k$ grows, the performance should increase. However, we notice that there is a decrease in performance when using $R_k$=50. This is directly related to the \textit{coarse-to-fine} hierarchical classification, in which at the coarse stage, we define how many species references will be taken to the fine stage using the genus reference candidates that appear in the ranking list $R_k$. 

\begin{figure}
	\includegraphics[width=.48\textwidth]{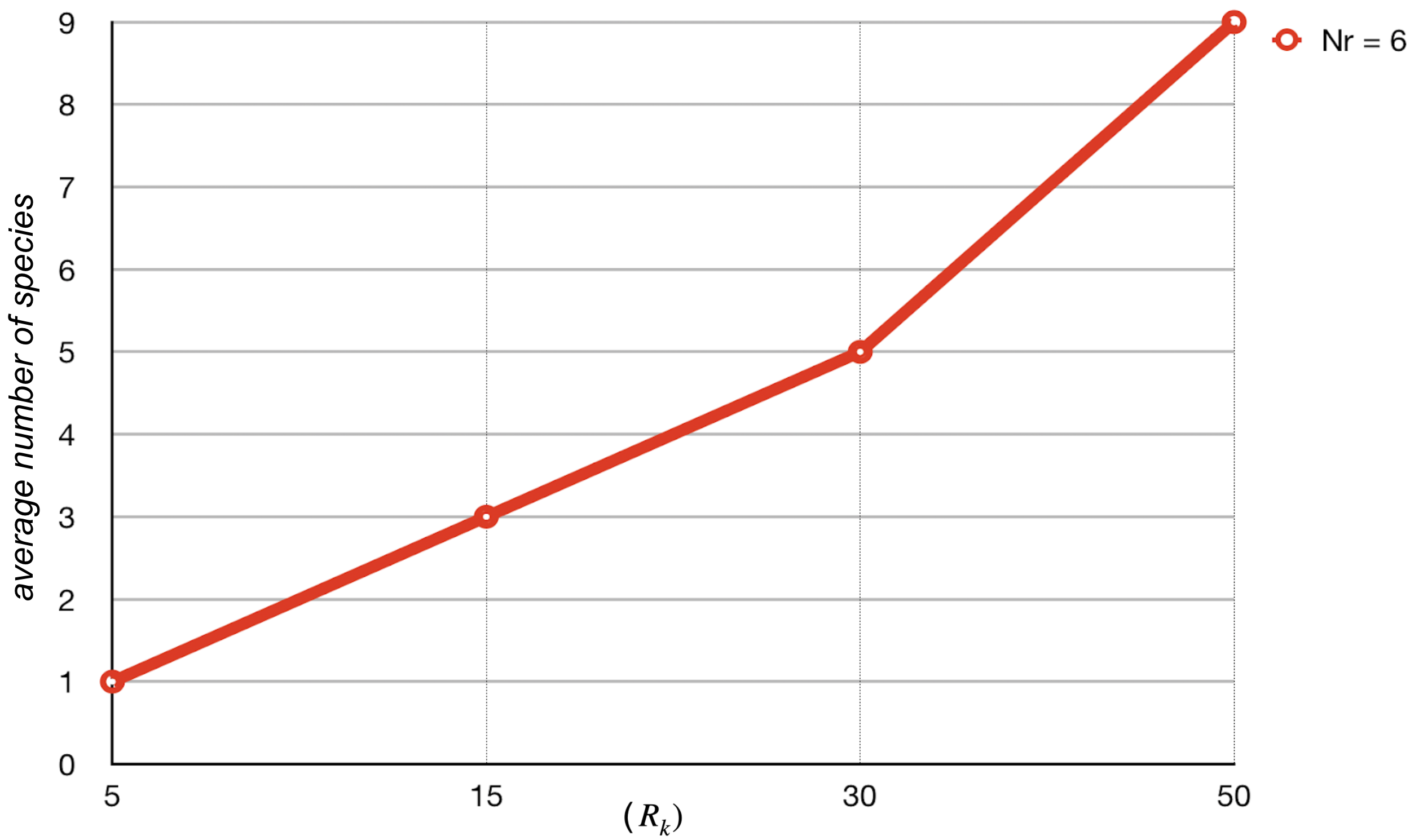}
\caption{Average number of species provided by the coarse stage, when using ($N_r$=6) concerning different sizes of genus references candidates returned in the ranking list ($R_k$)}.
\label{AverageSpecies}
\end{figure}

The idea behind Figure \ref{AverageSpecies} is to show the number of species (in average) that is sent to the second stage of the system when considering different sizes (k values) for the ranking list $R_k$ of genus candidates generated in the first system stage. The species selected are those that belong to the genus candidates in the list. However, the number of species may vary according to list size. For instance, with $N_r = 6$ (number of references), the average number of species processed in the second stage for $R_k = 50$, 30, and 15 is 9, 5, and 3, respectively. Such an analysis gave us some idea about decreasing the classification complexity in the second stage when we reduce the size of the list of genus candidates.

\begin{table}[width=.99\linewidth,pos=h]
\caption{Average classification score ($S$) of the S-CNN proposed method for the PlantCLEF 2015 dataset. 
}\label{resultsLifeCLEF}
\begin{tabular}{lcccc}
\toprule
Hierarchical Stage & $(R_k)$ & \multicolumn{3}{c}{Number of References ($N_r$)} \\
  &  &  1  &  3  &  6  \\ \midrule
1st (Genus)    & 5   & 0.72           & 0.77  & 0.77            \\
1st (Genus)    & 15   & 0.86           & 0.84  & 0.81            \\
\textbf{1st (Genus)}    & \textbf{30} & \textbf{0.98}   & \textbf{0.96}           & \textbf{0.95}              \\
1st (Genus)    & 50   & 0.99           & 0.98  & 0.98            \\

2nd (Species)  & 1    & 0.81           & 0.86  & 0.87            \\

\bottomrule
\multicolumn{5}{l}{$N_r$: number of reference samples per class in the}\\
\multicolumn{5}{l}{classification phase;} 

\end{tabular}
\end{table}

\begin{table}[width=.99\linewidth,pos=h]
\caption{Overall accuracy ($acc$) of the S-CNN proposed method for the LeafSnap dataset.
}\label{resultsLeafSnap}
\begin{tabular}{@{}lcccc@{}}
\toprule
Hierarchical Stage & $(R_k)$ & \multicolumn{3}{c}{Number of References ($N_r$)} \\
  &  &  1  &  3  &  6  \\ \midrule
1st (Genus)    & 5   & 0.91           & 0.92  & 0.95            \\
1st (Genus)    & 15   & 0.96           & 0.96  & 0.98            \\
\textbf{1st (Genus)}    & \textbf{30}   & \textbf{0.99}           & \textbf{0.98}  & \textbf{0.98}            \\
1st (Genus)    & 50   & 0.99           & 0.98  & 0.97                  \\
2nd (Species)  & 1    & 0.91           & 0.95  & 0.96            \\

\bottomrule
\multicolumn{5}{l}{$N_r$: number of reference samples per class in the}\\
\multicolumn{5}{l}{classification phase;}
\end{tabular}
\end{table}

\begin{table}[width=.99\linewidth,pos=h]
\caption{Final accuracy of the S-CNN proposed method considering $N_r$=6, $R_k$=30 in the first stage (genus) and top-$k$=1, 3 and 5 in the second stage (species). VGG16 baseline is comparable for each dataset using top-5 results}.\label{ResultsTopN}
\begin{tabular}{@{}lccccc@{}}
\toprule
Dataset & Method &Top-1 & Top-3 & Top-5 \\ \midrule
\multirow{2}{*}{PlantCLEF 2015} & S-CNN               & 0.87 & 0.94 &1.0           \\
  & VGG16               & 0.78 & 0.81 &0.85           \\
\multirow{2}{*}{LeafSnap} & S-CNN                  & 0.96    & 0.99       &1.0        \\ 
 & VGG16                 & 0.88    & 0.90       &0.93        \\ 
\bottomrule
\end{tabular}
\end{table}

\subsubsection{Analysis of the S-CNN proposed method}
\label{anali1}
\begin{figure}
	\includegraphics[width=.48\textwidth]{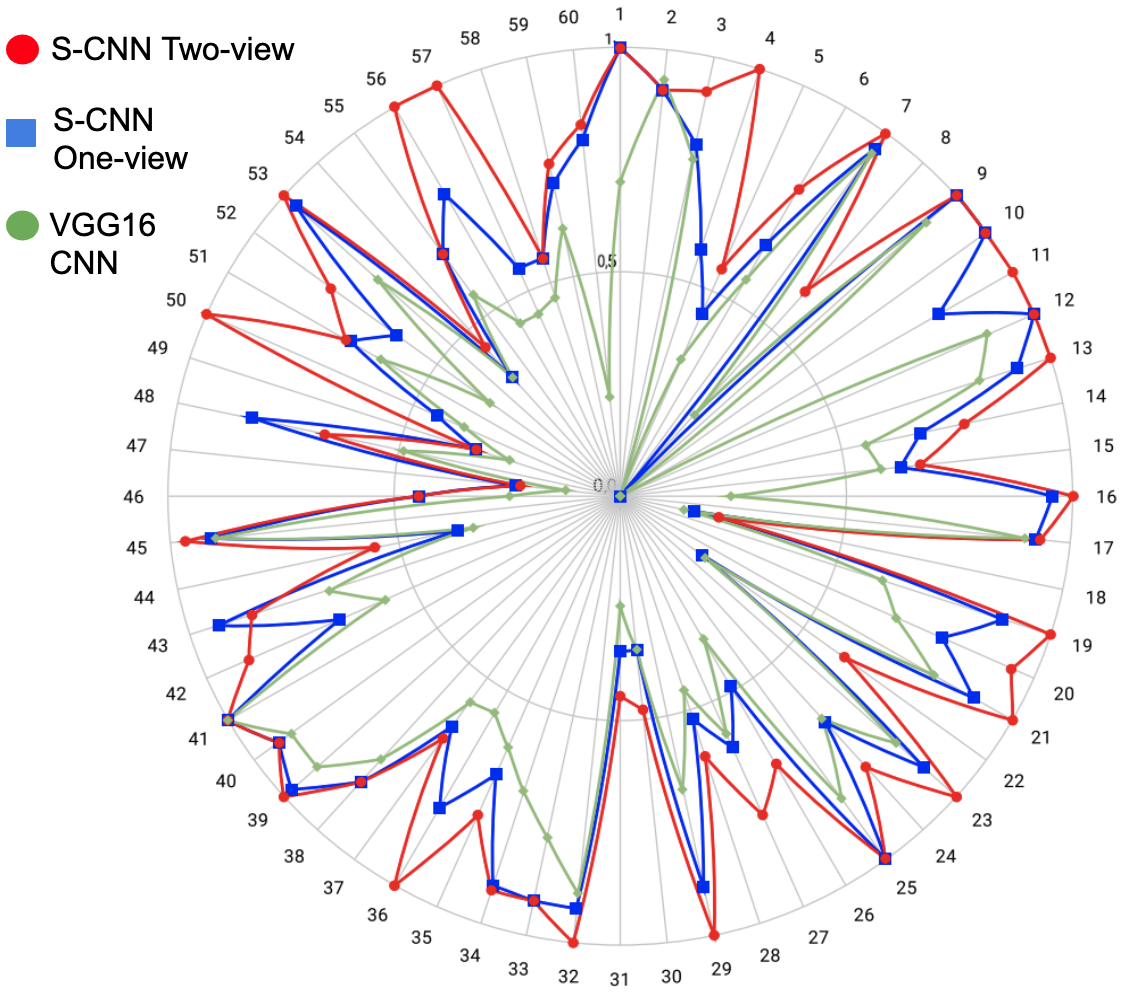}
\caption{Comparison of the proportion of correctly recognized leaves of each species at the coarse-to-fine classification using one-view (S-CNN one-view), the proposed method using two-view (S-CNN two-view) and pre-trained VGG16. 60 species are evaluated from PlantCLEF 2015 dataset.}
\label{radarLifeClef}
\end{figure}

\begin{figure}
	\centering
		\includegraphics[width=.48\textwidth]{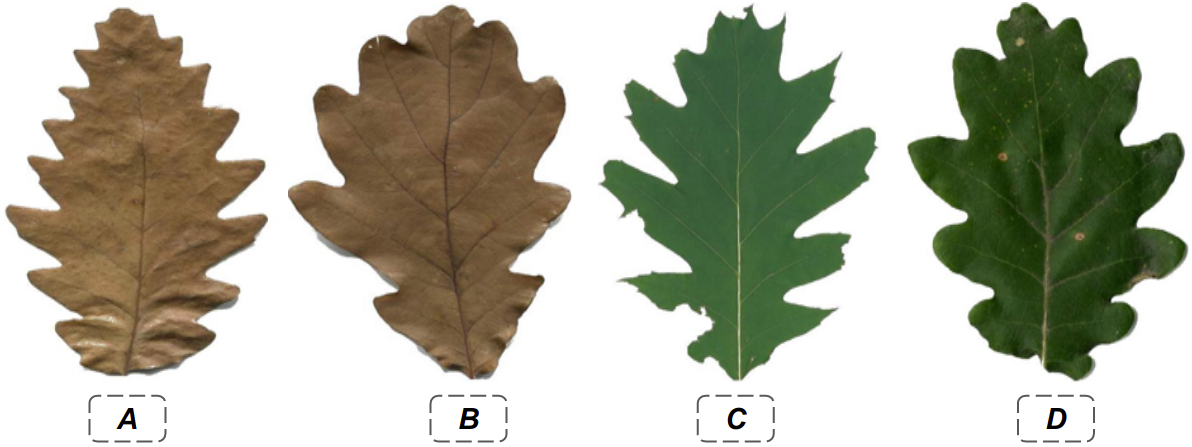}
	\caption{Samples of confused species (global representation): a)\textit{quercus cerris}; b)\textit{quercus petraea}; c)\textit{quercus rubra}; d)\textit{quercus pubescens}.}
	\label{similarSamples}
\end{figure}

Figure \ref{radarLifeClef} shows a spider web chart in which one can see the percentage of correctly recognized leaf images by each species considering the S-CNN proposed method using just one-view (global representation in both levels of hierarchy),  with two views (global and local representations) and the pre-trained VGG16 model. We considered the $S$ metric results on the PlantCLEF 2015 test dataset. According to Figure \ref{radarLifeClef}, the S-CNN two-view proposed method (red line) improves the classification rates of several species compared to the one-view purpose (blue line). The use of the local view applying cropped leaf images was suitable to reduce the conflict between species, corroborating with the preliminary experiment handled in Section \ref{anali2XXX}. For instance, the eighth species, named \textit{quercus cerris}, has 0.0 of accuracy when using just the one-view representation. On the other hand, it increased to 0.55 when we used the two-view proposed method. As we can see, when just one view is considered, the \textit{quercus cerris} is confused with \textit{quercus petraea}, \textit{quercus rubra}, and \textit{quercus pubescens}. Figure~\ref{similarSamples} shows the leaves \textit{quercus cerris}, \textit{quercus petraea}, \textit{quercus rubra}, and \textit{quercus pubescens} species respectively. Noticeably, the four species in the Figure~\ref{similarSamples} have similar morphological characteristics, which explains the confusion when just one view is adopted.

\begin{figure}
	\includegraphics[width=.48\textwidth]{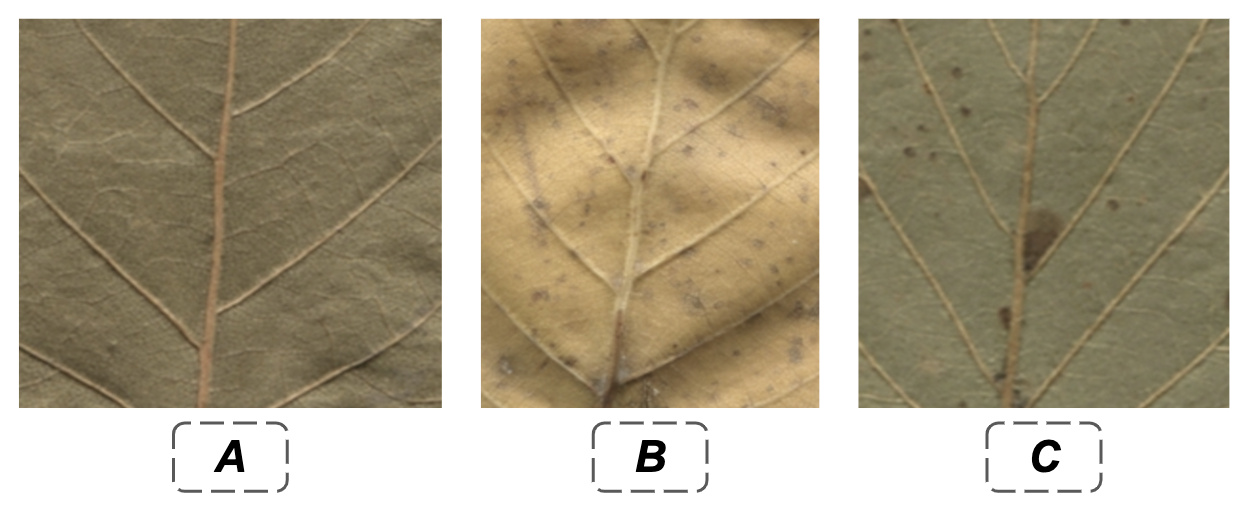}
\caption{Samples of confused species (local representation): a)\textit{betula pendula}; b)\textit{betula betulus} and c)\textit{betula avellana}.}
\label{similarCroppedSamples}
\end{figure}

However, it is important to observe that the accuracy has dropped off for few species using the S-CNN two-view proposed method. This is the case of the forty-eighth species in Figure \ref{radarLifeClef}. It is the \textit{betula pendula} specie. Observing the output of the two-view proposed method for related species, we found confusion between it and the \textit{betula betulus} and \textit{betula avellana}. They have very similar texture and vein patterns as shown in Figure \ref{similarCroppedSamples}.

Finally, we compare S-CNN approaches with the pre-trained VGG16 method presented in Section \ref{anali2XXX}. Table \ref{ComparisonApprochs} shows the final accuracies of VGG16, S-CNN two-view proposed method, and the S-CNN using one-view on the test set for both datasets: PlantCLEF 2015 and LeafSnap. The final results improved the recognition rates for the two-view S-CNN, in which we assure the effectiveness of the proposed method.

\begin{table}
\caption{Final accuracies for VGG16, S-CNN using one-view (global) and S-CNN proposed method for each dataset, PlantCLEF 2015 and LeafSnap.}\label{ComparisonApprochs}
\begin{tabular}{ccc}
\toprule
  & { PlantCLEF 2015 } & { LeafSnap } \\
 Approach &  ($S$)  &  ($acc$) \\\midrule
 VGG16 & 0.78 & 0.88 \\
 S-CNN one-view & 0.81 & 0.92 \\
 
 \textbf{S-CNN two-view}  & \multirow{2}{*}{\textbf{0.87} }  & \multirow{2}{*}{\textbf{0.96} } \\
 \textbf{proposed method} &  &  \\
\bottomrule
\end{tabular}
\end{table}

\subsubsection{\textbf{Impact of unbalanced data}
\label{exp2}}

Table \ref{GeralPlantCLEF2015} shows the impact of unbalanced data in the species accuracy when using the S-CNN proposed method and the pre-trained VGG16 model. As we can see, the second and third columns show the number of training images per species used for both models. It is important to say that the VGG16 model is trained using the original number of training samples shown in Table \ref{TableDefaultDatasets} for the PlantCLEF 2015 dataset, while the S-CNN model is trained using only six samples per class.

We observed that for almost all species, the performance of the S-CNN model is equal or better than the VGG16. Most of the mistakes of the VGG16 model occur in the species with few samples for training. As one may see, \textit{L. triphylla} species (bold line in Table \ref{GeralPlantCLEF2015}) has nine samples for training and seven samples for testing, it is most difficult for the VGG16 to generate an appropriate model to discriminate \textit{L. triphylla} among all species with so few samples for training. On the other hand, for the same case (\textit{L. triphylla} species), the S-CNN model trained with only 6 samples was able to recognize 6 out of 7 test samples correctly. The reason is that the S-CNN is not trained to learn a regular classifier of plant species but a distance metric to provide the similarity between two images (reference and test image). Therefore, the S-CNN model is capable of dealing with unbalanced data. Consequently, the S-CNN uses just six samples per class, and achieved a total of 182 hits. In contrast, the VGG16 hit 147 test samples with an unbalanced number of samples per class.

\addtolength{\tabcolsep}{-1pt}    
\begin{table*} 
\caption{Species and number of training and test images from the PlantCLEF 2015 dataset. Hits are the quantity of correct prediction per species held by the proposed S-CNN approach or VGG16.}\label{GeralPlantCLEF2015}
\begin{center}
\begin{tabular}{lccccc|lccccc} 
\hline
 & \multicolumn {2}{c}{Training}    &  & \multicolumn {2}{c}{Hits}  &  & \multicolumn {2}{c}{Training}    &  & \multicolumn {2}{c}{Hits}  \\
  Species &  VGG16 &  S-CNN & Test & VGG16  & S-CNN  & Species &  VGG16 &  S-CNN & Test & VGG16  & S-CNN \\ 
\hline
  V. opulus    & 60 &  6      & 9  & 8  & 8  & 
  M. papyrifera& 115   &  6   & 1   & 1 & 0     \\
    V. tinus    & 243 &  6     & 3  & 3 & 3   & M. carica   & 116   &  6   & 1   & 1 & 1     \\
   L. styraciflua    & 61  &   6    & 3  & 3 & 3  & M. rubra    & 88    &  6   & 1   & 1 & 1     \\
   A. prostrata   & 6   &   6    & 1   & 0 & 1  & Fra.~excelsior   & 60    &  6   & 9   & 4 & 5    \\
  A. cotinus       & 165 &   6    & 4   & 3 & 3  &  Fra.~ornus   & 73   &  6    & 1   & 1 & 1    \\
   A. pistacia       & 157   &   6  & 1  & 1 & 1  & Fra.~vahl    & 132    &  6  & 7   & 4 & 4       \\
  I. aquifolium     & 84   &  6& 1    & 1 & 1  & O. europaea    & 311   &  6   & 1    & 1 & 1    \\
   H. helix     & 264  &    6  & 9  & 5 & 6  &  S. bulgaris    & 101   &  6   & 3  & 2 & 3      \\
   R. aculeatus    & 290    & 6   & 25 & 22 & 23   &   P. hispanica    & 119    &  6  & 5   & 4 & 4       \\
 A. trichomanes   & 6     &  6  & 1   & 0 & 1   & C. monogyna    & 197   &  6   & 3  & 2 & 2    \\
    A. vulgaris     & 6     &  6   & 1   & 0 & 1 & C. germanica    & 50    &  6   & 1  & 1 & 1       \\
  B. glutinosa    & 51    &  6   & 2  & 2 & 2  & P. avium    & 67    &  6  & 2   & 2 & 2     \\
  B. pendula    & 122   &  6   & 6   & 2 & 4    & P. mahaleb    & 56   &   6   & 3  & 2 & 3     \\
  B. betulus    & 124    & 6   & 4  & 3 & 3    & P. spinosa    & 46    &  6   & 1 & 1 & 1     \\
    B. avellana    & 148   &   6  & 2   & 1 & 1  & U. minor   & 382    &  6   & 2  & 2 & 2       \\
  C. australis    & 190   &  6   & 4  & 1 & 1   &  Po. alba    & 197    &  6  & 2  & 1 & 2      \\
  F. cercis   & 142   &  6   & 2   & 2 & 2   &  Po. nigra    & 222   &  6   & 2  & 1 & 2     \\
  F. robinia    & 109   &  6   & 3   & 3 & 3  & Po. tremula    & 97   &  6    & 4  & 3 & 4      \\
 Q. cerris    & 125    &  6  & 9   & 2 & 5  & S. cinerea    & 24   &  6    & 2  & 2 & 2        \\
 Q. sylvatica   & 96     &  6  & 2   & 1 & 2  &  A. pseudo.      & 44    &  6   & 3  & 2 & 3      \\
   Q. pubescens   & 104   &  6   & 6   & 3 & 2   & A. sacchar.      & 42    &  6   & 5  & 2 & 4     \\
  Q. sativa    & 77     &  6  & 2   & 2 & 1  & A. negundo   & 111    &  6  & 1  & 1 & 1       \\
    Q. petraea    & 76   &  6    & 1 & 1 & 1    & A. platanoides   & 67     &  6  & 5  & 3 & 5        \\
   Q. rubra    & 48    &  6   & 2  & 1 & 2 &  A. campestre    & 160    &  6  & 13  & 9 & 12        \\
  G. genarium    & 19    &   6  & 1  & 1 & 1  & A. monspess    & 167   &  6   & 1 & 1 & 1       \\
  Gi. biloba    & 127    &   6 & 9   & 8 & 9  &    \textbf{L. triphylla}   & \textbf{9}     &  \textbf{6}   & \textbf{7}  &  \textbf{2} & \textbf{6}   \\
   L. nobilis    & 151   &   6  & 3  & 2 & 3  & A. altissima    & 82    &   6  & 4  & 1 & 3       \\
L. tulipifera    & 70    &  6   & 2  & 1 & 2    & T. baccata    & 10     &   6 & 2  & 1 & 2    \\
  T. tilia   & 71    &  6   & 4   & 2 & 3  & S. torminalis    & 36    &   6 & 3  & 3 & 3      \\
  T. cordata    & 28    &   6  & 3 &1 & 2  & B. davidii    & 126   &   6  & 1 & 1 & 1         \\
  
\hline
\end{tabular}
\end{center}
\end{table*}
\addtolength{\tabcolsep}{1pt}

We also performed experiments considering the unbalanced data on the LeafSnap dataset. Table \ref{UnbalancedLeafSnap} shows the use of different sizes of training samples, starting from 1, 3, 6, 10, 15, 25, and between 30 to 300 samples per species. For testing the LeafSnap dataset, we select randomly 15 samples by each class to compose the testing set. Concerning the first experiment (\#1) in Table \ref{UnbalancedLeafSnap}, it is difficult to get a good classifier in both cases since the S-CNN needs more than one pair of positive and negative samples to train and update the loss function to converge the model, while the VGG16 model requires a substantial number of training samples to provide solid results. The main advantage of the S-CNN is observed in the experiments \#2 and \#3, in which it outperforms widely the VGG16 using few training samples. 

{However, the S-CNN results in Table \ref{UnbalancedLeafSnap} tend to go down as long as we increase the number of training images. We can observed that the performance of the S-CNN starts to fall in the experiments \#4, \#5, \#6, and \#7. We believe that a huge quantity of image pairs (positive and negative) may cause the S-CNN overfitting, since most of their parameters are frozen when the fine-tuning is performed. We increase the number of positive and negative training samples substantially when we generate the pairwise to train the S-CNN considering experiments \#6 and \#7. For instance, using 25 (experiment \#6) samples for 184 species, we have a total of 55,200 positive pairs and 82,800 negative pairs.} 

Differently, the performance of the VGG16 starts to increase in the experiment \#6,  achieving 0.84 of accuracy ($acc$) when using 25 training samples per species. In experiment \#7 we have the best result for VGG16 with 0.88 of accuracy, when between 30 and 300 training samples were used. However, S-CNN outperforms the VGG16 in experiment \#3, reaching 0.96 of accuracy using just six training samples per class. These last experiments corroborate the ability of the S-CNN to deal with unbalanced data.

\begin{table}[width=.99\linewidth,cols=3,pos=h]
\caption{Accuracy achieved for different quantity of training images per class (data size) using S-CNN proposed method and VGG16 model for LeafSnap dataset. 
}\label{UnbalancedLeafSnap}
\begin{tabular}{@{}ccccccc@{}}


\toprule
\multirow{2}{*}{Experiment}     &  \multicolumn{2}{c}{Dataset Size}   &S-CNN & VGG16
\\ \cline{2-3}
     & Training  &  Testing & ($acc$) & ($acc$)\\
\midrule
     \#1 & 1     & 15    & 0.51&   0.53       \\
  \#2 & 3        & 15     & 0.85 &   0.52    \\
  \#3    & 6       & 15     & 0.96 &  0.53       \\ 
  \#4    & 10       & 15     & 0.94&  0.65       \\ 
  \#5    & 15       & 15     & 0.92&   0.78      \\ 
  \#6    & 25       & 15     & 0.91&   0.84      \\ 
  \#7    & [30-300]       & 15     & 0.88 & 0.88        \\ 
  \bottomrule
\end{tabular}
\end{table}

\subsubsection{\textbf{Stability and Scalability}
\label{exp3}}

Since the final result may be affected by the random selection of reference samples used to represent the plant species, we have evaluated the stability of the proposed method in such a situation. With that in mind, we performed five executions of the proposed method with different plant samples as references (without repetitions). We performed a data augmentation for species with few training samples using rotations in the original leaf image as presented in \cite{Araujo2018_2}. Table \ref{multipletimes} shows the results of the five executions for the PlantCLEF 2015 and LeafSnap test sets. As one can see, the method performance is stable even using different sets of references.

\begin{table}
\caption{Executions with distinct sets of reference images, six references are randomly selected to compose the reference sets for each dataset, PlantCLEF 2015 and LeafSnap. Av = average performance.}\label{multipletimes}
\begin{tabular}{lcccccc}
\toprule
Dataset &  \multicolumn{5}{c}{Execution} \\
\cline{2-6}
  &  1&  2  &  3  &  4  &  5 & \textbf{Av} \\ \midrule
PlantCLEF 2015  & 0.87 & 0.86 & 0.84  & 0.85  & 0.87& \textbf{0.86}\\
LeafSnap        & 0.96 & 0.94  & 0.93  & 0.95  & 0.93& \textbf{0.94}\\
\bottomrule
\end{tabular}
\end{table}

The scalability of the proposed method can be evaluated by considering plant species unseen during the training step. For such an aim, in the proposed method, it is just necessary to add reference images of the new species to be considered. Therefore, one of the proposed method's main advantages is that it does not require retraining the S-CNNs, avoiding such a time-consuming process.

\begin{figure}
\centering
	\includegraphics[width=.48\textwidth]{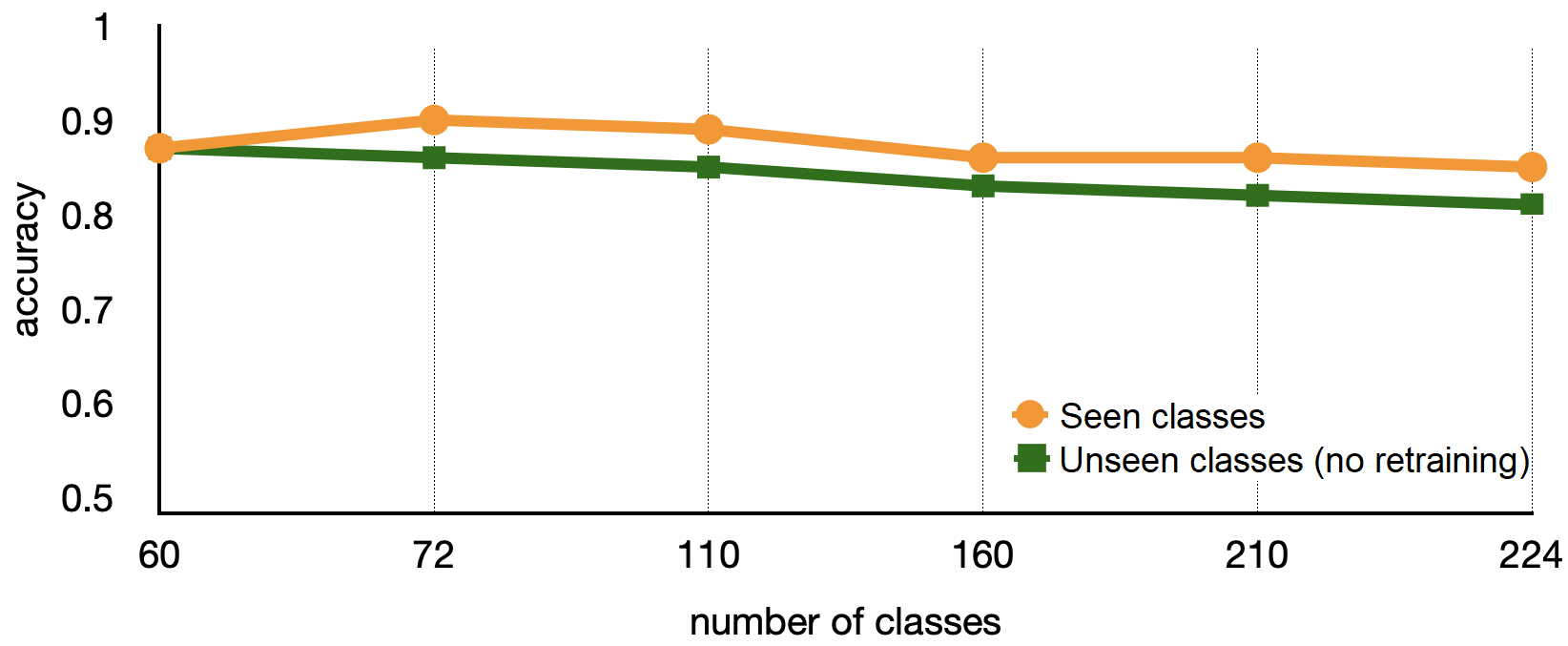}
\caption{System scalability considering unseen and seen leaf species.}
\label{scalabilityClass}
\end{figure}

We experimentally evaluated the impact on the system recognition performance by considering incrementally new plant species not seen during training. For this purpose, we prepared small subsets with 12, 50, 100, 150, and 184 plant species using the LeafSnap dataset. For each new subset added, we computed the performance of the system originally trained on the PlantCLEF 2015 dataset (60 species), with and without a retraining process. As expected, we always observed the best performance with the system retraining. However, we noticed a small loss in performance (up to 0.4) when considering unseen species, as shown in Figure~\ref{scalabilityClass}.  As we can see, the accuracy before considering unseen classes was 0.87 for the 60 classes of the PlantCLEF 2015 dataset. After adding 12, 50, 100, 150, and 184 new species, which belong to the LeafSnap dataset, the accuracy dropped to 0.86, 0.85, 0.83, 0.82, and 0.81, respectively. It is important to note that even adding 184 unseen species (new ones) from another different dataset, the proposed method sustained an accuracy close to that achieved for only 60 species. This behavior is impressive since the model never saw these new species during training. The S-CNNs can compute the similarity between the new references and the test images. With these results, we can say that the proposed method scales relatively well. Besides, they also indicate that the system performed well in the case of a cross dataset evaluation.

Table~\ref{TableTime} shows the necessary time to classify a single leaf image for a growing number of species. For calculating the computational time, we look at just the test phase. We observed that the computational time grows slower than the linear function as the number of reference images grows. 

\begin{table}

\caption{Computational time for classify one leaf plant considering the scalability of classes. 
}\label{TableTime}
\begin{tabular}{@{}llc@{}}
\toprule
Number of & &Time \\ 
Classes & Dataset & (sec) \\ 
\midrule
60 & PlantCLEF 2015 & 0.2010 \\
72 & PlantCLEF 2015 + LeafSnap & 0.3965 \\
110 & PlantCLEF 2015 + LeafSnap & 0.9276 \\
160 & PlantCLEF 2015 + LeafSnap & 1.3830 \\
210 & PlantCLEF 2015 + LeafSnap & 1.6789 \\
244 & PlantCLEF 2015 + LeafSnap & 1.8458 \\
\bottomrule
\end{tabular}

\end{table}

\subsubsection{\textbf{Comparison with the State-of-the-Art}
\label{exp4}}

We compared the result achieved by the proposed method with those achieved by related works, which have also used either PlantCLEF 2015 or LeafSnap datasets. As we can see in Table \ref{comparativeTable}, in a previous work \cite{Araujo2018_2}, we reported almost the same results with a CNN approach on PlantCLEF 2015 dataset. It worth mentioning that there are no significant differences regarding the accuracy. However, the novel approach proposed here works better in unbalanced data scenarios, using the S-CNN architecture trained on few samples, which avoid a substantial computational cost of using a data augmentation strategy as previously employed in \cite{Araujo2018_2}. Another important improvement is that the proposed method is easily scalable, and unknown plant species can be easily integrated into the S-CNN models without retraining them.

\addtolength{\tabcolsep}{-2pt}
\begin{table}
\caption{Comparison with the state-of-the-art for leaf classification for PlantCLEF 2015 and LeafSnap datasets.}\label{comparativeTable}
\begin{tabular}{rccc}
\toprule
&  & {\scriptsize PlantCLEF 2015 } & {\scriptsize LeafSnap } \\
Reference & Approach & {\scriptsize ($S$) } & {\scriptsize ($acc$)} \\\midrule
\citet{Sungbin2015} & CNN & 0.76 & - \\
\citet{Lee17HGOCNN} & CNN & 0.80 & - \\
\citet{Araujo2018_2} & CNN & 0.86 & - \\
\citet{Ghazi2017} & CNN & 0.84 & - \\
\citet{spatialstructureSiamese} & S-CNN & 0.84 & - \\
\citet{Bodhwani2019} & CNN &  - & 0.93 \\
{\color{black}\citet{Song2019}} & CNN &  - & 0.91 \\
{\color{black}\citet{Hu2018}} & CNN &  - & 0.85 \\
{\color{black}\citet{Riaz2020}} & CNN &  - & 0.99 \\
\bf Proposed Method & \bf S-CNN & \bf 0.87 & \bf 0.96 \\
\bottomrule
\end{tabular}
\end{table}

\addtolength{\tabcolsep}{2pt}

{
Eight out of nine related works present in Table \ref{comparativeTable} use CNN models \cite{Sungbin2015, Lee17HGOCNN, Ghazi2017, Araujo2018_2, Bodhwani2019, Song2019, Hu2018, Riaz2020}, while just one use S-CNN model \cite{spatialstructureSiamese}. In \cite{spatialstructureSiamese} to address the fine-grained problem, a piece of additional information is used considering multi-views of the plant organs like a leaf, entire plant, and flowers.  Another interesting work based on S-CNN, but not included in Table \ref{comparativeTable} is that described in \cite{Wang2019}.  The authors have shown the power of S-CNNs by using just twenty (20) training samples per class on the LeafSnap dataset, but they have used just ten (10) plant species for testing. \citet{Riaz2020} achieved the best results for the LeafSnap dataset (field subset) using a multi-level representation strategy. However, they considered more than one viewing angle, combining different plant organs. Besides, they performed data augmentation to make the dataset balanced.} 
 
Considering all these methods, the proposed strategy of using S-CNN, a two-view leaf image representation, and a coarse-to-fine classification has shown to be a promising approach. It provides a competitive performance using a small subset of training samples while producing a scalable solution.
 
\section{Conclusion}
\label{sec5}

We proposed a novel method based on a two-view leaf image representation and hierarchical classification strategy for fine-grained plant species recognition. The botanical taxonomy is used to drive a coarse-to-fine classification strategy to identify the plant genus and species. The two-view representation of a plant leaf can improve recognition performance using global (shape and color) and local features (texture and plant veins). A deep metric based on an S-CNN was used to reduce the dependence of the proposed method on a large amount of training samples. Besides that, the S-CNN makes the proposed method scalable and unknown plant species can be easily integrated into the S-CNN models without the need of retraining them.

The experiments on two challenging fine-grained datasets of leaf images (PlantCLEF 2015 and LeafSnap) confirmed the effectiveness of the  proposed method – the recognition accuracy over those two datasets reaches 0.87 and 0.96, respectively. As future work, we plan to deal with auto-encoders to learn representations inside an S-CNN architecture with hierarchical property for leaf plant recognition.

\bibliographystyle{cas-model2-names}

\balance


\newpage

\end{document}